\documentclass[a4paper]{article}

\usepackage[english]{babel}
\usepackage[utf8]{inputenc}
\usepackage[tbtags]{amsmath}
\usepackage[round]{natbib}
\usepackage{graphicx}
\usepackage{caption}
\usepackage{subcaption}
\usepackage{amsfonts}
\usepackage{dsfont}
\usepackage{bm}
\usepackage{verbatim}
\usepackage[colorinlistoftodos]{todonotes}
\usepackage{authblk}

\usepackage{mathtools}
\usepackage{a4wide}

\usepackage[linesnumbered]{algorithm2e}
\usepackage{algorithmic}

\newcommand{\RR}{\mathbb{R}}
\newcommand{\EE}{\mathbb{E}}

\newcommand{\SG}{\operatorname{SG}}

\newcommand{\bt}[1]{{\mathbf{#1}}}
\newcommand{\bth}{{\boldsymbol{\theta}}}
\newcommand{\bepsi}{{\boldsymbol{\epsilon}}}
\newcommand{\eqdef}{\overset{\text{def}}{=}}
\usepackage[symbol]{footmisc}

\newcommand{\norm}[1]{\left\lVert#1\right\rVert}

\usepackage{lineno}

\makeatletter
\DeclareFontEncoding{LS1}{}{}
\DeclareFontSubstitution{LS1}{stix}{m}{n}
\DeclareMathAlphabet{\mathscr}{LS1}{stixscr}{m}{n}
\makeatother

\title{Biologically inspired alternatives to backpropagation through time for learning in recurrent neural nets}

\author[*]{Guillaume Bellec}
\author[*]{Franz Scherr}
\author[ ]{Elias Hajek}
\author[ ]{Darjan Salaj}
\author[ ]{Robert Legenstein}
\author[ ]{Wolfgang Maass}
\affil[ ]{Institute for Theoretical Computer Science, Graz University of Technology, Austria}

\date{\today}

\begin{document}
\maketitle

*First authors

\abstract{The way how recurrently connected networks of spiking neurons in the brain acquire powerful information processing capabilities through learning has remained a mystery. This lack of understanding is linked to a lack of learning algorithms for recurrent networks of spiking neurons (RSNNs) that are both functionally powerful and can be implemented by known biological mechanisms. Since RSNNs are simultaneously a primary target for implementations of brain-inspired circuits in neuromorphic hardware, this lack of algorithmic insight also hinders technological progress in that area. The gold standard for learning in recurrent neural networks in machine learning is back-propagation through time (BPTT), which implements stochastic gradient descent with regard to a given loss function. But BPTT is unrealistic from a biological perspective, since it requires a transmission of error signals backwards in time and in space, i.e., from post- to presynaptic neurons. We show that an online merging of locally available information during a computation with suitable top-down learning signals in real-time provides highly capable approximations to BPTT. For tasks where information on errors arises only late during a network computation, we enrich locally available information through feedforward eligibility traces of synapses that can easily be computed in an online manner. The resulting new generation of learning algorithms for recurrent neural networks provides a new understanding of network learning in the brain that can be tested experimentally. In addition, these algorithms provide efficient methods for on-chip training of RSNNs in neuromorphic hardware.

We changed in this version 2 of the paper the name of the new learning algorithms to \textit{e-prop}, corrected minor errors, added details -- especially for resulting new rules for synaptic plasticity, edited the notation, and included new results for TIMIT.}



\section*{Introduction}

A characteristic property of networks of neurons in the brain is that they are recurrently connected: „the brain is essentially a multitude of superimposed and ever-growing loops between the input from the environment and the brain’s outputs“ \citep{buzsaki2006rhythms}. In fact, already \citep{lorente1938architectonics} had proposed that synaptic loops were the basic circuits of the central nervous system, and a large body of experimental work supports this view \citep{kandel2000principles}. Recurrent loops of synaptic connections occur both locally within a lamina of a cortical microcircuit, between their laminae, between patches of neural tissue within the same brain area, and between different brain areas. Hence the architecture of neural networks in the brain is fundamentally different from that of feedforward deep neural network models that have gained high attention because of their astounding capability in machine learning \citep{lecun2015deep}.

Recurrently connected neural networks tend to provide functionally superior neural network architectures for tasks that involve a temporal dimension, such as video prediction, gesture recognition, speech recognition, or motor control. Since the brain has to solve similar tasks, 
and even transforms image recognition into a temporal task via eye-movements, 
there is a clear functional reason why the brain employs recurrently connected neural networks. In addition, recurrent networks enable the brain to engage memory on several temporal scales, and to represent and continuously update internal states as well as goals. Furthermore the brain is a powerful prediction machine that learns through self-supervised learning to predict the consequences of its actions and of external events. In fact, predictions provide the brain with a powerful strategy for compensating the relative slowness of its sensory feedback.

The computational function of recurrently connected neural networks in the brain arises from a combination of nature and nurture that has remained opaque. In particular, it has remained a mystery how recurrent networks of spiking neurons (RSNNs) can learn. Recurrent networks of artificial neurons are commonly trained in machine learning through BPTT. BPTT can not only be used to implement supervised learning, but -- with a suitably defined loss function $E$ -- self-supervised, unsupervised and reward based learning. Unfortunately BPTT requires a physically unrealistic propagation of error signals backwards in time. This feature 
also thwarts an efficient implementation in neuromorphic hardware. It even hinders an efficient implementation of BP or BPTT on GPUs and other standard computing hardware: „backpropagation results in locking -- the weights of a network module can only be updated after a full forward propagation of data, followed by loss evaluation, and then finally after waiting for the backpropagation of error gradients“ \citep{czarnecki2017understanding}. Locking is an issue of particular relevance for applications of BPTT to recurrent neural networks, since this amounts to applications of backpropagation to the unrolled recurrent network, which easily becomes several thousands of layers deep.

We show that BPTT can be represented by a sum of products based on a new factorization or errors gradients with regards to the synaptic weights $\theta_{ji}$.
The error gradient is represented here as a sum over $t$ of an eligibility trace $e_{ji}^t$ until time $t$ - which is independent from error signals - and a learning signal ${L}_j^t$ that reaches this synapse at time $t$, see equation \eqref{eq:grad}.
This can be interpreted as on online merging for every time step $t$ of eligibility traces and learning signals.
Because of the prominent role which forward propagation of eligibility traces play in the resulting approximations to BPTT we refer to these new algorithms as \textbf{\textit{e-prop}}.

The key problem for achieving good learning results with eligibility traces is the online production of suitable learning signals that gate 
the update of the synaptic weight at time $t$. In order to achieve the full learning power of BPTT, this learning signal would still have to be complex and questionable from a biological perspective. But several biologically plausible approximations of such online learning signals turn out to work surprisingly well, especially for tasks that recurrent networks of neurons in the brain need to solve.


There exists an abundance of experimental data on learning- or error signals in the brain. A rich literature documents the error-related negativity (ERN) that is recorded by EEG-electrodes from the human brain. The ERN has the form of a sharp negative-going deflection that accompanies behavioral errors, for example in motor control. Remarkable is that the ERN appears very fast, even before direct evidence of a behavioral error becomes accessible through sensory feedback (see e.g. Fig. 4 in \citep{maclean2015using}), suggesting that it employs an internal error prediction network. Furthermore the amplitude of the ERN correlates with improved performance on subsequent trials (\citep{gehring1993neural}, see also the review in \citep{buzzell2017development}). These results suggest that the ERN is in fact a signal that gates learning. The data of \citep{buzzell2017development} also shows that the ERN is generated by a distributed system of brain areas, in which posterior cingulate cortex, dorsal anterior cingulate, and parietal cortex assume dominant roles from early stages of development on. Furthermore, error-related activity from additional brain areas -- insula, orbitofrontal cortex, and inferior frontal gyrus -- increases with age. These experimental data suggest 
that error signals in the human brain are partially innate, but are complemented and refined during development.

The precise way how these error signals gate synaptic plasticity in the brain is unknown. One conjectured mechanism involves top-down disinhibition of dendrites and neurons, e.g. by activating VIP-interneurons in layer 1, which inhibit somatostatin-positive (SOM+) inhibitory neurons. Hence the activation of VIP neurons temporarily removes the inhibitory lock which SOM+ neurons hold on activity and plasticity in distal dendrites of pyramidal cells 
\citep{pi2013cortical}. 
Another pathway for top-down regulation of synaptic plasticity involves cholinergic activation of astrocytes \citep{sugihara2016cell-specific}. Furthermore the cerebellum is known to play a prominent role in the processing of error signals and gating plasticity \citep{d'angelo2016distributed} Most importantly, the neuromodulator dopamine plays an essential role in the control of learning, in particular also for learning of motor skills \citep{hosp2011dopaminergic}.
Experimental data verify that neuromodulators interact with local eligibility traces in gating synaptic plasticity, see \citep{gerstner2018eligibility} for a review. Of interest for the context of this paper is also the recent discovery that dopaminergic neurons in the mid-brain do not emit a uniform global signal, but rather a multitude of spatially organized signals for different populations of neurons \citep{engelhard2018specialized}. This distributed architecture of the error-monitoring system in the brain is consistent with the assumption that local populations of neurons receive different learning signals that have been shaped during evolution and development.

Error signals in the brain are from the functional perspective reminiscent of error signals that have turned out to alleviate the need for backprogation of error signals 
in feedforward neural networks. A particularly interesting version of such signals is called broadcast alignment (BA) in \citep{samadi2017deep} and direct feedback alignment in \citep{nokland2016direct}. These error signals are sent directly from the output stage of the network to each layer of the feedforward network. 
If one applies this broadcast alignment idea to 
the unrolled feedforward version of a recurrent network, one still runs into the problem that an error broadcast to an earlier time-slice or layer would have to go backwards in time.
We present a simple method where this can be avoided, which we call \textit{e-prop 1}.

Besides BA we explore in this paper two other methods for generating learning signals that provide -- in combination with eligibility traces -- powerful alternatives to BPTT. In \textit{e-prop 2} we apply the Learning-to-Learn (L2L) framework to train separate neural networks  -- called error modules -- to produce suitable learning signals for large families of learning tasks. But in contrast to the L2L approach of \citep{wang2016learning} and \citep{duan2016rl} we allow the recurrent neural network to modify its synaptic weights for learning a particular task. Only the synaptic weights within the error module are determined on the larger time scale of the outer loop of L2L (see the scheme in Figure \ref{fig:one-shot}).
We show that this approach opens new doors for learning in recurrent networks of spiking neurons, enabling for example one-shot learning of pattern generation. 
Our third method, \textit{e-prop 3}, employs the synthetic gradient approach of \citep{jaderberg2016decoupled} and \citep{czarnecki2017understanding}. We show that eligibility traces substantially enhance the power of synthetic gradients, surpassing in some cases even the performance of full BPTT for artificial neural networks. Altogether the \textit{e-prop} approach suggests that a rich reservoir of algorithmic improvements of network learning waits to be discovered, where one employs dedicated modules and processes for generating learning signals that enable learning without backpropagated error signals. In addition this research is likely to throw light on the functional role of the complex distributed architecture of brain areas that are involved in the generation of learning signals in the brain.

These \textit{e-prop} algorithms have an attractive feature from the theoretical perspective: They can be viewed -- and analyzed -- as approximations to a theoretically ideal: stochastic gradient descent, or BPTT. \textit{E-prop} algorithms are also of particular interest from the perspective of understanding learning in RSNNs of the brain. They tend to provide better learning performance for RSNNs than previously known methods. In addition, in contrast to most of the previously used methods, they do not require biologically unrealistic ingredients. In fact, it turns out that network learning with \textit{e-prop} provides a novel understanding of refined STDP rules \citep{clopath2010connectivity} from a network learning perspective, that had been proposed in order to fit detailed experimental data on local synaptic plasticity mechanisms \citep{ngezahayo2000synaptic,sjostrom2001rate,nevian2006spine}.

In addition, \textit{e-prop} provides a promising new approach for implementing on-chip learning in RSNNs that are implemented in neuromorphic hardware, such as Brainscales \citep{schemmel2010wafer}, SpiNNaker \citep{furber2014spinnaker} and Loihi \citep{davies2018loihi}. Backpropagation of error signals in time as well as locking are formidable obstacles for an efficient implementation of BPTT on a neuromorphic chip. These obstacles are alleviated by the \textit{e-prop} method.

Synaptic plasticity algorithms involving eligibility traces and gating factors have been reviewed for reinforcement learning 
in \citep{fremaux2016neuromodulated}, see \citep{gerstner2018eligibility} for their relationships to data. We re-define eligibility traces for the different context of gradient descent learning. 
Eligibility traces are used in classical reinforcement learning theory \citep{sutton1998introduction} 
to relate the (recent) history of network activity to later rewards. This reinforcement learning theory inspired our approach on a conceptual level, but the details of the mathematical analysis become quite different, since we address a different problem: how to approximate error gradients in recurrent neural networks.

We will derive in the first section of Results the basic factorization equation \eqref{eq:grad} that underlies our \textit{e-prop} approach. We then discuss applications of \textit{e-prop 1} to RSNNs with a simple BA-like learning signal. In the subsequent section we will show how an application of L2L in \textit{e-prop 2} can improve the learning capability of RSNNs. Finally, we show in the last subsection on \textit{e-prop 3} that adding eligibility traces to the synthetic gradient approach of \citep{jaderberg2016decoupled} and \citep{czarnecki2017understanding}  improves learning also for recurrent networks of artificial neurons.

\section*{Results}

\begin{figure}
    \centering
    \includegraphics[width=\textwidth]{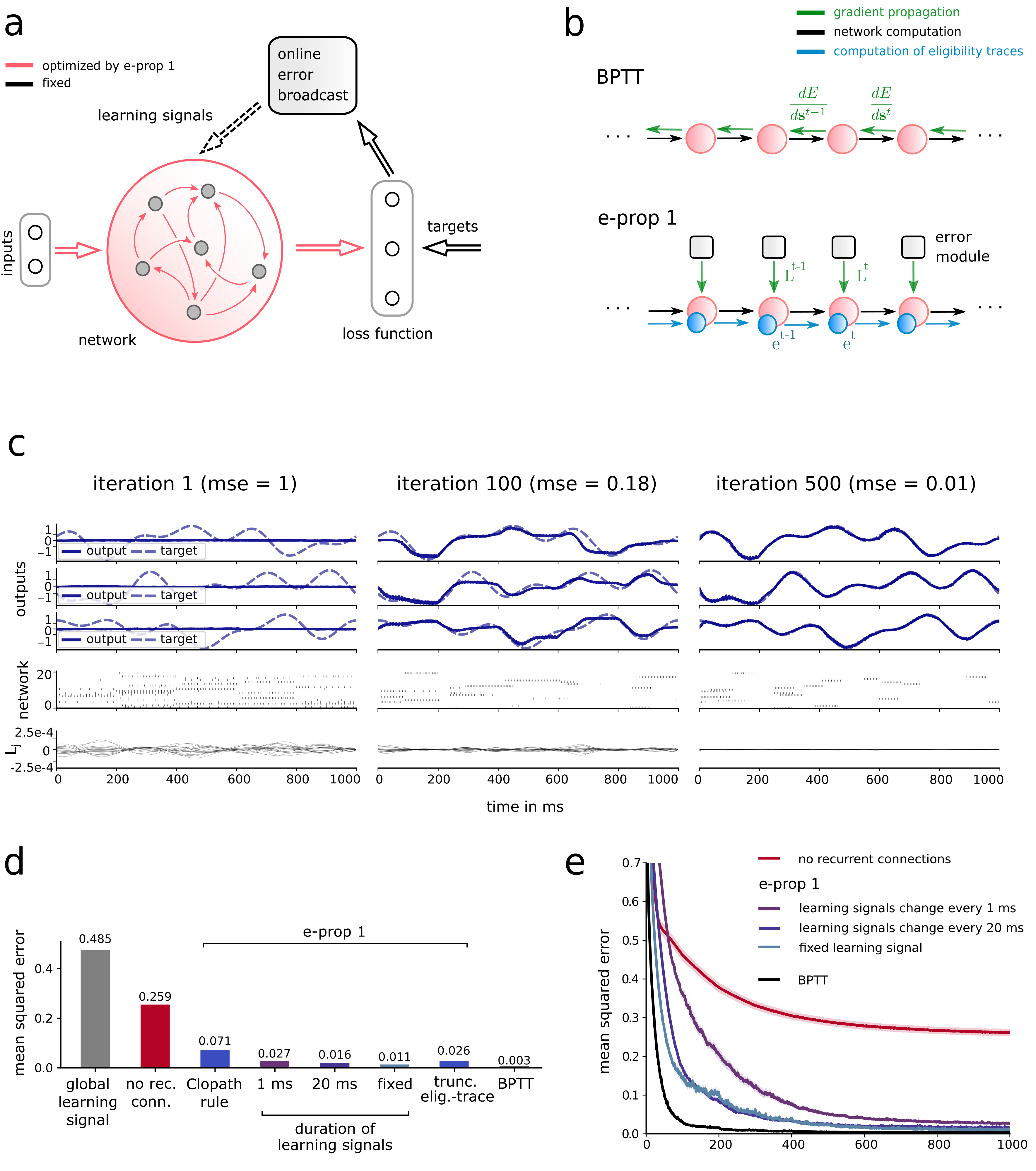}
    \caption{ \textbf{Scheme and performance of \textit{e-prop 1}} \textbf{a})~Learning architecture for \textit{e-prop 1}. The error module at the top sends online error signals with random weights to the network that learns.
    	\textbf{b})~Temporal dynamics of information flows in BPTT and \textit{e-prop} algorithms. The propagation of error signals backwards in time of BPTT is replaced in \textit{e-prop} algorithms by an additional computation that runs forward in time: the computation of eligibility traces.
    	\textbf{c})~Evaluation of \textit{e-prop 1} for a classical benchmark task for learning in recurrent SNNs: Learning to generate a target pattern, extended here to the challenge to simultaneously learn to generate 3 different patterns, which makes credit assignment for errors more difficult. 
    	\textbf{d})~Mean squared error of 
    	several learning algorithms for this task. “Clopath rule'' denotes a replacement of the resulting synaptic plasticity rule of \textit{e-prop 1} by the rule proposed in \citep{clopath2010connectivity} based on experimental data.
    	\textbf{e})~Evolution of the mean squared error during learning. 
    	}	
    \label{fig:partials}
\end{figure}


\paragraph{Network models}
The learning rules that we describe can be applied to a variety of recurrent neural network models:
Standard RSNNs consisting of leaky integrate-and-fire (LIF) neurons, LSNNs (Long short term memory Spiking Neural Networks) that also contain adaptive spiking neurons \citep{bellec2018long}, and networks of LSTM (long short-term memory) units \citep{hochreiter_long_1997}.
LSNN were introduced to capture parts of the function of LSTM network in biologically motivated neural network models.
In order to elucidate the link to biology, we focus in the first two variants of \textit{e-prop} on the LIF neuron model (see Figures \ref{fig:partials}, \ref{fig:store_recall} and \ref{fig:one-shot}).
The LIF model is a simplified model of biological neurons: each neuron integrates incoming currents into its membrane potential, and as soon as the membrane potential crosses a threshold from below, the neuron ``spikes'' and a current is sent to subsequent neurons. Mathematically, the membrane potential is a leaky integrator of a weighted sum of the input currents, and the spike is a binary variable that becomes non-zero when a spike occurs (see equation \eqref{eq:lifv} and \eqref{eq:lifz} in Methods). To enrich the temporal processing capability of the network (see Figure \ref{fig:store_recall}),
a portion of the neurons in an LSNN have adaptive firing thresholds. The dynamics of the adaptive thresholds is defined in equation \eqref{eq:alifb} in Methods.
We also applied a third variant of \textit{e-prop 3} to LSTM networks to show that \textit{e-prop} algorithms can be competitive on machine learning benchmarks (see Figure \ref{fig:copy}).

To describe the common core of these \textit{e-prop} algorithms, all network models are subsumed under a general formalism.
We assume that each neuron $j$ is at time $t$ in an internal state $\bt s_j^t \in \mathbb{R}^d$ and emits an observable state $z_j^t$.
We also assume that $\bt s_j^t$ depends on the other neurons only through the vector $\bt z^{t-1}$ of observable states of all neurons in the network. Then, the network dynamics takes for some functions $M$ and $f$ the form: $\bt s_j^{t} = M(\bt s_j^{t-1}, \bt z^{t-1}, \bt x^t, \bth)$ and $z_j^t=f(\bt s_j^t)$, where $\bth$ is the vector of model parameters (in the models considered here, synaptic weights).
For instance for LIF neurons with adaptive thresholds, the internal state $\bt s_j^t$ of neuron $j$ is a vector of size $d=2$ formed by the membrane voltage and the adaptive firing threshold, and the observable state $z_j^t \in \{0, 1\}$ indicates whether the neuron spikes at time $t$. The definition of the functions $M$ and $f$ defining the neuron dynamics for this model are given by equations \eqref{eq:lifv},\eqref{eq:lifz} and \eqref{eq:alifb} in Methods and illustrated in Figure \ref{fig:graph}.

\paragraph{Mathematical framework for \textit{e-prop} algorithms}
The fundamental mathematical law that enables the \textit{e-prop} approach is that the gradients of BPTT can be factorized as a sum of products between learning signals $L_j^t$ and eligibility traces $e_{ji}^t$. 
We subsume here under the term eligibility trace that information which is locally available at a synapse and does not depend on network performance.
The online learning signals $L_j^t$ are provided externally and could for example quantify how spiking at the current time influences current and future errors.
The general goal is to approximate the gradients of the network error function $E$ with respect to the model parameters $\bt{\theta}_{ji}$. If the error function $E$ depends exclusively on the network spikes $E(\mathbf{z}^1, \dots, \mathbf{z}^T)$, the fundamental observation for \textit{e-prop} is that the gradient with respect to the weights can be factorized as follows (see Methods for a proof):
\begin{equation}
\frac{d E}{d \theta_{ji}} = \sum_{t} L_j^t ~ e_{ji}^t~. \label{eq:grad}
\end{equation}

We refer to $L_j^t$ and $e_{ji}^t$ as the learning signals and eligibility traces respectively, see below for a definition. 
Note that we use distinct notation for the partial derivative $\frac{\partial E(\mathbf{z}_1, \dots, \mathbf{z}_T)}{\partial \mathbf{z}_1}$, which is the derivative of the mathematical function $E$ with respect to its first variable $\mathbf{z}_1$, and the total derivative $\frac{d E(\mathbf{z}_1, \dots, \mathbf{z}_T)}{d \mathbf{z}_1}$ which also takes into account how $E$ depends on $\bt z_1$ indirectly through the other variables $\mathbf{z}_1, \dots, \mathbf{z}_T$. The essential term for gradient descent learning is the total derivative $\frac{d E}{d \theta_{ji}}$. It has usually been computed with BPTT \citep{werbos1990backpropagation} or RTRL \citep{williams1989learning}.

\paragraph{Eligibility traces}
The intuition for eligibility traces is 
that a synapse remembers some of its activation history while ignoring inter neuron dependencies.
Since the network dynamics is formalized through the equation $\bt s_j^{t} = M(\bt s_j^{t-1}, \bt z^{t-1}, \bt x^t, \bth)$, the internal neuron dynamics isolated from the rest of the network is described by $D_j^{t-1} = \frac{\partial M}{\partial \bt s_j^{t-1}}(\bt s_j^{t-1}, \bt z^{t-1}, \bt x^t, \bth)  \in \mathbb{R}^{d \times d}$ (recall that $d$, is the dimension of internal state of a single neuron; $d=1$ or $2$ in this paper). Considering the partial derivative of the state with respect to the synaptic weight $ \frac{\partial M}{\partial \bth_{ji}}(\bt s_j^{t-1}, \bt z^{t-1}, \bt x^t, \bth) \in \RR^{d}$ (written $\frac{\partial \bt s_j^{t}}{\partial \theta_{ji}}$ for simplicity), we formalize the mechanism that retains information about the previous activity at the synapse $i \rightarrow j$ by the eligibility vector $\bt{\bepsi}_{ji}^{t} \in \RR^d$ defined with the following iterative formula:
\begin{eqnarray}
\bt{\bepsi}_{ji}^{t} & = & D_j^{t-1} \cdot \bt{\bepsi}_{ji}^{t-1} + \frac{\partial \bt s_j^{t}}{\partial \theta_{ji}}~, \label{eq:elig-vector}
\end{eqnarray}
where $\cdot$ is the dot product. Finally, this lead to the eligibility trace which is the scalar product between this vector and the derivative $\frac{\partial f}{\partial \bt s_j^t} (\bt s_j^t)$ (denoted $\frac{\partial z_j^t}{\partial \textbf{s}_j^t}$ for simplicity) which captures how the existence of a spike $z_j^t$ depends on the neuron state $\bt s_j^t$:
\begin{eqnarray}
e_{ji}^{t} & = & \frac{\partial z_j^t}{\partial \textbf{s}_j^t} \cdot \bt{\bepsi}_{ji}^{t}~.  \label{eq:elig-scalar}
\end{eqnarray}
%

In practice for LIF neurons the derivative $\frac{\partial z_j^t}{\partial \bt s_j^t}$ is ill-defined due to the discontinuous nature of spiking neurons. As done in \citep{bellec2018long} for BPTT, this derivative is replaced in simulations by a simple nonlinear function of the membrane voltage $h_j^t$ that we call the pseudo-derivative (see Methods for details).
The resulting eligibility traces $e_{ji}^t$ for LIF neurons are the product of a the post synaptic pseudo derivative $h_j^t$ with the trace $\hat{z}_i^t$ of the presynpatic spikes (see equation \eqref{eq:elig-vector-LIF} in Methods).
For adaptive neurons in LSNNs and for LSTM units the computation of eligibility traces becomes less trivial, see equations \eqref{eq:elig-vector-ALIF} and \eqref{eq:elig-scalar-ALIF}. But they can still be computed in an online manner forward in time, along with the network computation. We show later that the additional term arising for LSNNs in the presence of threshold adaptation holds a crucial role when working memory has to be engaged in the tasks to be learnt.

Note that in RTRL for networks of rate-based (sigmoidal) neurons \citep{williams1989learning}, the error gradients are computed forward in time by multiplying the full Jacobian $\bt J$ of the network dynamics with the tensor $\frac{d \bt s_{k}^{t}}{d \theta_{ji}}$ that computes the dependency of the state variables with respect to the parameters: $\frac{d \bt s_{k}^{t}}{d \theta_{ji}}=\sum_{k'}~\bt J_{kk'}^t~\cdot~\frac{d \bt s_{k'}^{t-1}}{d \theta_{ji}}~+~\frac{\partial \bt s_k^{t}}{\partial \theta_{ji}}$ (see equation (12) in \citep{williams1989learning}). Denoting  with $n$ the number of neurons, this requires $O(n^4)$ multiplications, which is computationally prohibitive in simulations, whereas BPTT or network simulation requires only  $O(n^2)$ multiplications.
In \textit{e-prop}, the eligibility traces are $n \times n$ matrices which is one order smaller than the tensor $\frac{d \bt s_{k}^{t}}{d \theta_{ji}}$, also $D_j^t$ are $d \times d$ matrices which are restrictions of the full Jacobian $\bt J$ to the neuron specific dynamics. As a consequence, only $O(n^2)$ multiplications are required in the forward propagation of eligibility traces, their computation is not more costly than BPTT or simply simulating the network.

\paragraph{Theoretically ideal learning signals}
To satisfy equation \eqref{eq:grad}, the learning signals $L_j^t \in  \mathbb{R}^d$ can be defined by the following formula:
\begin{equation}
L_j^t \overset{\text{def}}{=} \frac{dE}{d z_j^t}~. \label{eq:L}
\end{equation}
Recall that $\frac{dE}{d z_j^t}$ is a total derivative and quantifies how much a current change of spiking activity might influence future errors.
As a consequence, a direct computation of the term $L_j^t$ needs to back-propagate gradients from the future as in BPTT.
However we show that \textit{e-prop} tends to work well if the ideal term $L_j^t$ is replaced by an online approximation $\widehat{L}_j^t$.
In the following three sections we consider three concrete approximation methods, that define three variants of \textit{e-prop}. If not clearly stated otherwise, all the resulting gradient estimations described below can be computed online and depend only on quantities accessible within the neuron $j$ or the synapse $i \rightarrow j$ at the current time $t$.

\paragraph{Synaptic plasticity rules that emerge from this approach}

The learning algorithms \textit{e-prop 1} and \textit{e-prop 2} that will be discussed in the following are for networks of spiking neurons. Resulting local learning rules are very similar to previously proposed and experimentally supported synaptic plasticity rules. They have the general form (learning signal) $\times$ (postsynaptic term) $\times$ (presynaptic term) as previously proposed 3-factor learning rules \citep{fremaux2016neuromodulated,gerstner2018eligibility}.
The general form is given in equation \eqref{eq:elig-scalar-LIF}, where $h_j^t$ denotes a postsynaptic term and the last factor $\hat{z}_i^{t-1}$  denotes the presynaptic term \eqref{eq:elig-vector-LIF}. These last two terms are similar to the corresponding terms in the plasticity rule of \citep{clopath2010connectivity}. It is shown in Figure \ref{fig:partials}d that one gets very similar results if one replaces the rule \eqref{eq:grad-LIF} that emerges from our approach by the Clopath rule.

The version \eqref{eq:e-prop1} of this plasticity rule for \textit{e-prop 1} contains the specific learning signal that arises in broadcast alignment as first factor.  For synaptic plasticity of adapting neurons in LSNNs the last term, the eligibility trace, becomes a bit more complex because it accumulates information over a longer time span, see equation \eqref{eq:elig-scalar-ALIF}. The resulting synaptic plasticity rules for LSTM networks are given by equation \eqref{eq:elig-scalar-lstm}.

\subsection*{\textit{E-prop 1}: Learning signals that arise from broadcast alignment}
A breakthrough result for learning in feedforward deep neural networks was the discovery that a substantial portion of the learning power of backprop can be captured if the backpropagation of error signals through chains of layers is replaced by layer specific direct error broadcasts, that consists of a random weighted sum of the errors that are caused by the network outputs; typically in the last layer of the network \citep{samadi2017deep,nokland2016direct}.
This heuristic can in principle also be applied to the unrolled version of a recurrent neural network, yielding a different error broadcast for each layer of the unrolled network, or equivalently, for each time-slice of the computation in the recurrent network. This heuristic would suggest to send to each time slice error broadcasts that employ different random weight matrices. We found that the best results can be achieved if
one chooses the same random projection of output errors for each time slice (see Figure~\ref{fig:partials}d and e).


\textbf{Definition of \textit{e-prop 1}:} \textit{E-prop 1} defines a learning signal that only considers the instantaneous error of network outputs and ignores the influence of the current activity on future errors.
As justified in Methods, this means that the approximation of the learning signal $\widehat{L}_j^t$ is defined by replacing the total error derivative $\frac{dE}{d z_j^t}$ with the partial derivative $\frac{\partial E}{\partial z_j^t}$.
Crucially, this replacement makes it possible to compute the learning signal in real-time, whereas the total derivative needs information about future errors which should be back-propagated through time for an exact computation.

To exhibit an equation that summarizes the resulting weight update, we consider a network of LIF neurons and output neurons formalized by $k$ leaky readout neurons $y_k^{t}$ with decay constant $\kappa$. If $E$ is defined as the squared error between the readouts $y_k^t$ and their targets $y_k^{*,t}$, and the weight updates are implemented with gradient descent and learning rate $\eta$, this yields (the proof and more general formula \eqref{eq:e-prop1-methods} are given in Methods):
\begin{eqnarray}
\Delta \theta_{ji}^{\mathrm{rec}} & = \eta & \sum_t \big( \sum_{k} \theta_{kj}^{\mathrm{out}} (y_k^{*,{t}}-y_k^{t}) \big)  \sum_{t' \leq t}\kappa^{t - t'} h_j^{t'} \hat{z}_{i}^{t'-1}~, \label{eq:e-prop1}
\end{eqnarray}
where $h_j^{t'}$ is a function of the post-synaptic membrane voltage (the pseudo-derivative, see Methods) and $\hat{z}_{i}^{t'}$ is a trace of preceding pre-synaptic spikes with a decay constant $\alpha$. This is a three-factor learning rule of a type that is commonly used to model experimental data \citep{gerstner2018eligibility}. However a less common feature is that, instead of a single global error signal, learning signals are neuron-specific weighted sums of different signed error signals arising from different output neurons $k$. For a more complex neuron model such as LIF with adaptive thresholds, which are decisive for tasks involving working memory \citep{bellec2018long}, the eligibility traces are given in equation \eqref{eq:elig-scalar-ALIF} and the learning rule in equation \eqref{eq:e-prop1-methods}. 

To model biology, a natural assumption is that 
synaptic connections to and from readout neurons are realized through different neurons. 
Therefore it is hard to conceive that the feedback weights are exactly symmetric to the readout weights, as required for gradient descent according to equation \eqref{eq:e-prop1} that follows from the theory. Broadcast alignment \citep{lillicrap2016random} suggests the replacement of $\theta_{kj}^{\mathrm{out}}$ by a random feedback matrix. We make the same choice to define a learning rule \eqref{eq:e-prop1-methods-LIF} with mild assumptions: the learning signal is a neuron-specific random projection of signed error signals $y_k^{*,{t}}-y_k^{t}$. Hence we replaced the weights $\theta_{kj}^{\mathrm{out}}$ with random feedback weights denoted $B_{jk}^{\mathrm{random}}$.
Other authors \citep{zenke2018superspike, kaiser2018synaptic} have derived learning rules similar to \eqref{eq:e-prop1-methods-LIF} for feedforward networks of spiking neurons but not for recurrent ones.
We test the learning performance of \textit{e-prop 1} for two generic computational tasks for recurrent neural networks: generation of a temporal pattern, and storing selected information in working memory.

\paragraph{Pattern generation task 1.1}
Pattern generation is an important component of motor systems. We asked whether the simple learning setup of \textit{e-prop 1} endows RSNNs with the capability to learn to generate patterns in a supervised manner.   

\textbf{Task:} To this end, we considered a pattern generation task which is an extension of the task used in \citep{nicola2017supervised}. 
In this task, the network should autonomously generate a three-dimensional target signal for 1 s. Each dimension of the target signal is given by the sum of four sinusoids with random phases and amplitudes. Similar to \citep{nicola2017supervised}, the network received a clock input that indicates the current phase of the pattern (see Methods).

\textbf{Implementation:}
The network consisted of 600 recurrently connected LIF neurons. All neurons in this RSNN projected to three linear readout neurons. All input, recurrent and output weights were plastic, see Figure \ref{fig:partials}a. A single learning trial, consisted of a 1 s simulation where the network produced a three-dimensional output pattern and gradients were computed using \textit{e-prop 1}. Network weights were updated after each learning trial (see Methods for details). 


\textbf{Performance:} Figure \ref{fig:partials}c shows the spiking activity of a randomly chosen subset of 20 of 600 neurons in the RSNN along with the output of the three readout neurons after application of \textit{e-prop 1} for 1, 100 and 500 seconds, respectively. In this representative example, the network achieved a very good fit to the target signal (normalized mean squared error 0.01). Panel d shows the averaged mean squared errors (mse) for several variants of \textit{e-prop 1} and a few other learning methods.

As an attempt to bridge a gap between phenomenological measurements of synaptic plasticity and functional learning models, we addressed the question whether a synaptic plasticity rule that was fitted to data in \citep{clopath2010connectivity} could reproduce the function of the second and third factors in equation \eqref{eq:e-prop1}. These two terms ($\hat{z}_i^{t-1}$ and $h_j^{t}$) couple the presynaptic and postsynaptic activity in a multiplicative manner. In the model of long term potentiation fitted to data by \citep{clopath2010connectivity}, the presynaptic term is identical but the postsynaptic term includes an additional non-linear factor depending on a filtered version of the membrane voltage.
We found that a replacement of the plasticity rule of \textit{e-prop 1} by the Clopath rule had little impact on the result, as shown in Figure \ref{fig:partials}d under the name ``Clopath rule'' (see equation  \eqref{eq:e-prop1-clopath} in Methods for a precise definition of this learning rule). 
We also asked whether these pre- and postsynaptic factors could be simplified further. When replacing the trace $\hat{z}_i^{t-1}$ and $h_j^{t}$ by the binary variable $z_i^{t-1}$, this just caused an increase of the mse from $0.011$ to $0.026$.
In comparison, when the network has no recurrent connections or when the learning signal is replaced by a uniform global learning signal ($B_{jk}=\frac{1}{\sqrt{n}}$ with $n$ the number of neurons) the mse increased by one order of magnitude to $0.259$ and $0.485$, respectively. This indicated the importance of diverse learning signals and recurrent connections in this task.

Panel e shows that a straightforward application of BA to the unrolled RSNNs, with new random matrices for error broadcast at every ms, or every 20 ms, worked less well or converged slower.
Finally, while \textit{e-prop 1} managed to solve the task very well (Figure \ref{fig:partials}d),
BPTT achieved an even lower mean squared error (black line in Figure \ref{fig:partials}e).

\begin{figure}
	\centering
	\includegraphics[width=\textwidth]{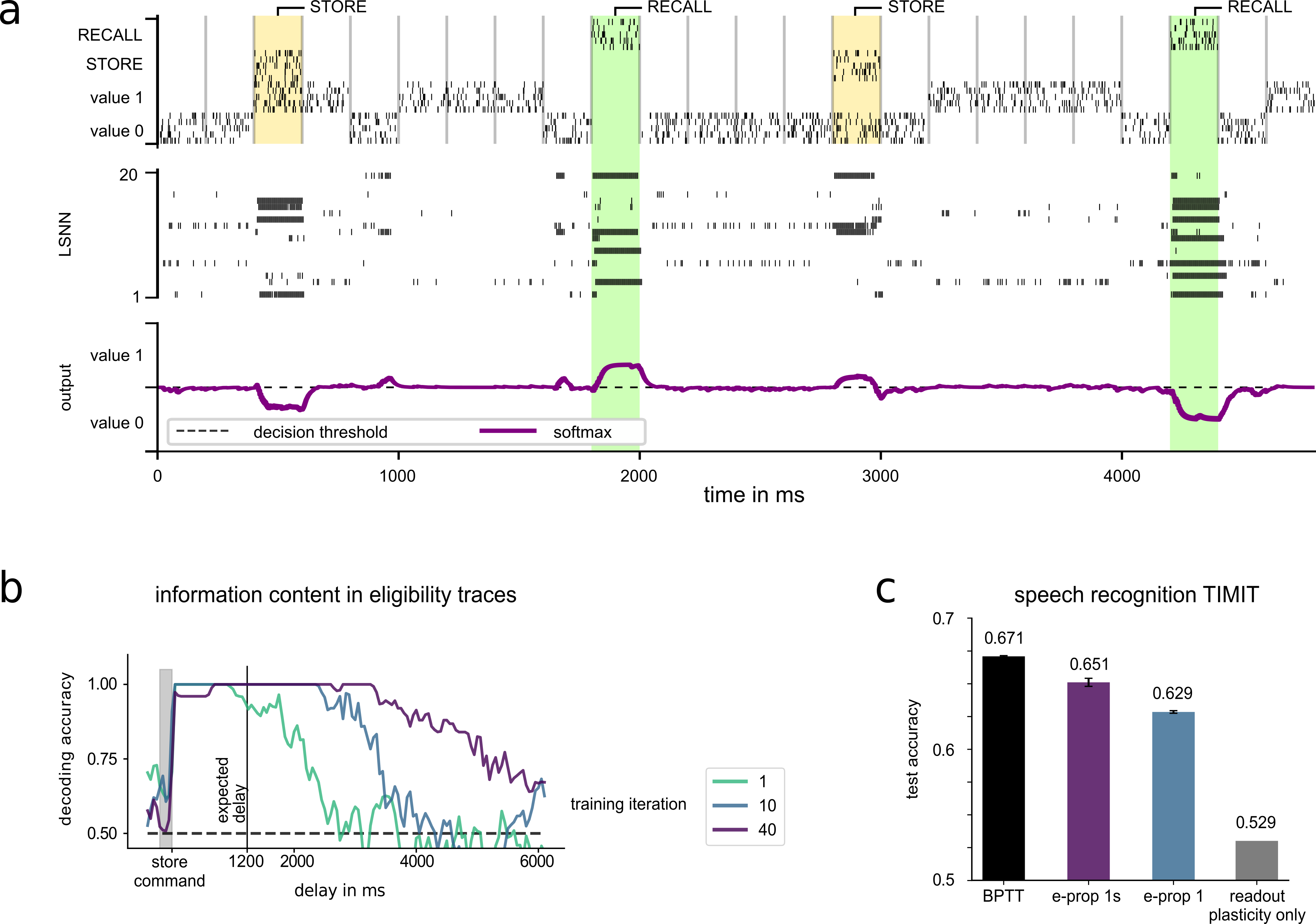}
	\caption{ \textbf{Testing \textit{e-prop 1} on the store-recall and speech recognition tasks.}
	\textbf{a})~The store-recall task requires to store the value 0 or 1 that is currently provided by other input neurons when a STORE command is given, and to recall it when a RECALL command is given. 
	It is solved with \textit{e-prop 1} for an LSNN. 
	\textbf{b})~Information content of eligibility traces (estimated via a linear classifier) about the bit that was to be stored during store-recall task. 
	\textbf{c})~Training LSNNs with \textit{e-prop 1} to solve the speech reccognition task on TIMIT dataset.
	\textit{e-prop 1s} indicates the case when the feedback weights are exactly symmetric to the readout weights.}
	\label{fig:store_recall}
\end{figure}

\paragraph{Store-recall task 1.2}
Many learning tasks for brains involve some form of working memory.
We therefore took a simple working memory task and asked whether \textit{e-prop 1} enables an LSNN to learn this task, in spite 
of a substantial delay between the network decision to store information and the time when the network output makes an error.

\textbf{Task:} The network received a sequence of binary values encoded by alternating activity of two groups of input neurons (“Value 0'' and “Value 1'' in Figure \ref{fig:store_recall}a, top). In addition, it received command inputs, STORE and RECALL, encoded likewise by dedicated groups of input neurons. The task for the network was to output upon a RECALL command the value that was present in the input at the time of the most recent STORE command. In other words, the network had to store a bit (at a STORE command) and recall it at a RECALL command. After a STORE, a RECALL instructions was given during each subsequent time period of length $D=200$ ms with probability $p_{\text{command}}=\frac{1}{6}$. This resulted in an expected delay of $\frac{D}{p_{\text{command}}}=1.2$ s between STORE and RECALL instruction. The next STORE appeared in each subsequent period of length $D$ with probability $p_\text{command}$. We considered the task as solved when the misclassification rate on the validation set reached a value below $5\%$.

\textbf{Implementation:}
We used a recurrent LSNN network consisting of $10$ standard LIF neurons and $10$ LIF neurons with adaptive thresholds. 
The input neurons marked in blue and red at the top of Figure \ref{fig:store_recall}a produced Poisson spike trains with time varying rates. An input bit to the network was encoded by spiking activity at $50$ Hz in the corresponding input channels for a time period of $D=200$ ms. STORE and RECALL instructions were analogously encoded through firing of populations of other Poisson input neurons at $50$ Hz. Otherwise input neurons were silent. The four groups of input channels consisted of $25$ neurons each.
During a recall cue, two readouts are competing to report the network output.
The readout with highest mean activation $y_k^t$ during the recall period determines which bit value is reported.
To train the network, the error function $E$ is defined as the cross entropy between the target bit values and the softmax of the readout activations (see Methods for details).
Importantly, an error signal is provided only during the duration of a RECALL command and informs about the desired change of output in that delayed time period.
The network was trained with \textit{e-prop 1}. Parameters were updated every $2.4$ s during training.

\textbf{Performance:}
Figure \ref{fig:store_recall}a shows a network trained with \textit{e-prop 1} that stores and recalls a bit accurately. This network reached a misclassification rate on separate validation runs below $5$\% in $50$ iterations on average with \textit{e-prop 1}. For comparison, the same accuracy was reached within $28$ iterations on average with full BPTT. We found that adaptive neurons are essential for these learning procedures to succeed:
A network of 20 non-adapting LIF neurons could not solve this task, even if it was trained with BPTT.

It might appear surprising that \textit{e-prop 1} is able to train an LSNN for this task, since the learning signal is only non-zero during a RECALL command. 
This appears to be problematic, because in order to reduce errors the network has to learn to handle information from the input stream in a suitable manner during a much earlier time window: during a STORE command, that appeared on average $1200$ ms earlier.
We hypothesized that this was made possible because eligibility traces can hold information during this delay.
In this way a learning signal could contribute to the modification of the weight of a synapse that had been activated much earlier, for example during a STORE command.
According to the theory (equation \eqref{eq:elig-scalar-ALIF} in the methods), eligibility traces of adapting neurons decay with a time constant comparable to that of the threshold adaptation. To verify experimentally that this mechanism makes it possible to hold the relevant information, we first verified that the same LSNN network failed at learning the task when eligibility traces are truncated by setting $\bt \bepsi_{ji}^t = \frac{\partial \bt s^t}{\partial \theta_{ji}}$. Second, we quantified the amount of information about the bit to be stored that is contained in the true eligibility traces. This information was estimated via the decoding accuracy of linear classifiers, and the results are reported in Figure~\ref{fig:store_recall}b.
While the neuron adaptation time constants were set to $1.2$~s, we found that the decoding accuracy quickly rises above chance even for much longer delays.
After $200$ training iterations, the relevant bit can be decoded up to $4$~s after the store signal arrived with high accuracy ($>90\%$ of the trials).

Altogether the results in Figure \ref{fig:partials} and \ref{fig:store_recall} suggest that \textit{e-prop 1} enables RSNNs to learn the most fundamental tasks which RSNNs have to carry out in the brain: to generate desired temporal patterns and to carry out computations that involve a working memory. Pattern generation tasks were also used for testing the performance of FORCE training for RSNNs in \citep{nicola2017supervised}. While FORCE training has not been argued to be biologically plausible because of the use of a non-local plasticity rule and the restriction of plasticity to readout neurons, \textit{e-prop 1} only engages mechanisms that are viewed to be biologically plausible. Hence it provides a concrete hypothesis how recurrent networks of neurons in the brain can learn to solve the most fundamental tasks which such networks are conjectured to carry out. More generally, we conjecture that \textit{e-prop 1} can solve all learning tasks that have been demonstrated to be solvable by FORCE training.
However we do not want to claim that \textit{e-prop 1} can solve all learning tasks for RSNNs that can be solved by BPTT according to \citep{bellec2018long}. But the power of \textit{e-prop} algorithms can be substantially enhanced by using more sophisticated learning signals than just random linear combinations of signed errors as in broadcast alignment. Several neural systems in the brain receive raw error signals from the periphery and output highly processed learning cues for individual brain areas and populations of neurons.
We propose that such neural systems have been refined by evolution and development to produce learning signals that enable more powerful versions of \textit{e-prop} algorithms, such as the ones that we will discuss in the following.

For implementations of \textit{e-prop} algorithms in neuromorphic hardware -- in order to enable efficient on-chip learning of practically relevant tasks --  another reservoir of mechanisms becomes of interest that also can make use of concrete aspects of specific neuromorphic hardware. For example, in order to approximate the performance of BPTT for training LSNNs for speech recognition, a test on the popular benchmark dataset TIMIT suggests that the accuracy of \textit{e-prop 1} $(0.629)$ can be brought closer to that achieved by BPTT ($0.671$; see \citep{bellec2018long}) by simply using the current values of readout weights, rather than random weights, for broadcasting current error signals (we label this \textit{e-prop 1} version \textit{e-prop 1s}). This yields an accuracy of $0.651$.
As a baseline for recurrent spiking neural networks we include a result for randomly initialized LSNN where only
the readout weights are optimized (see Figure~\ref{fig:store_recall}c readout plasticity only, accuracy $0.529$).
For comparison, the best result achieved by recurrent artificial neural networks (consisting of the most complex form of LSTM units) after extensive hyperparameter search  was $0.704$ \citep{greff2017lstm}.

\begin{figure}
	\centering
	\includegraphics[width=\textwidth]{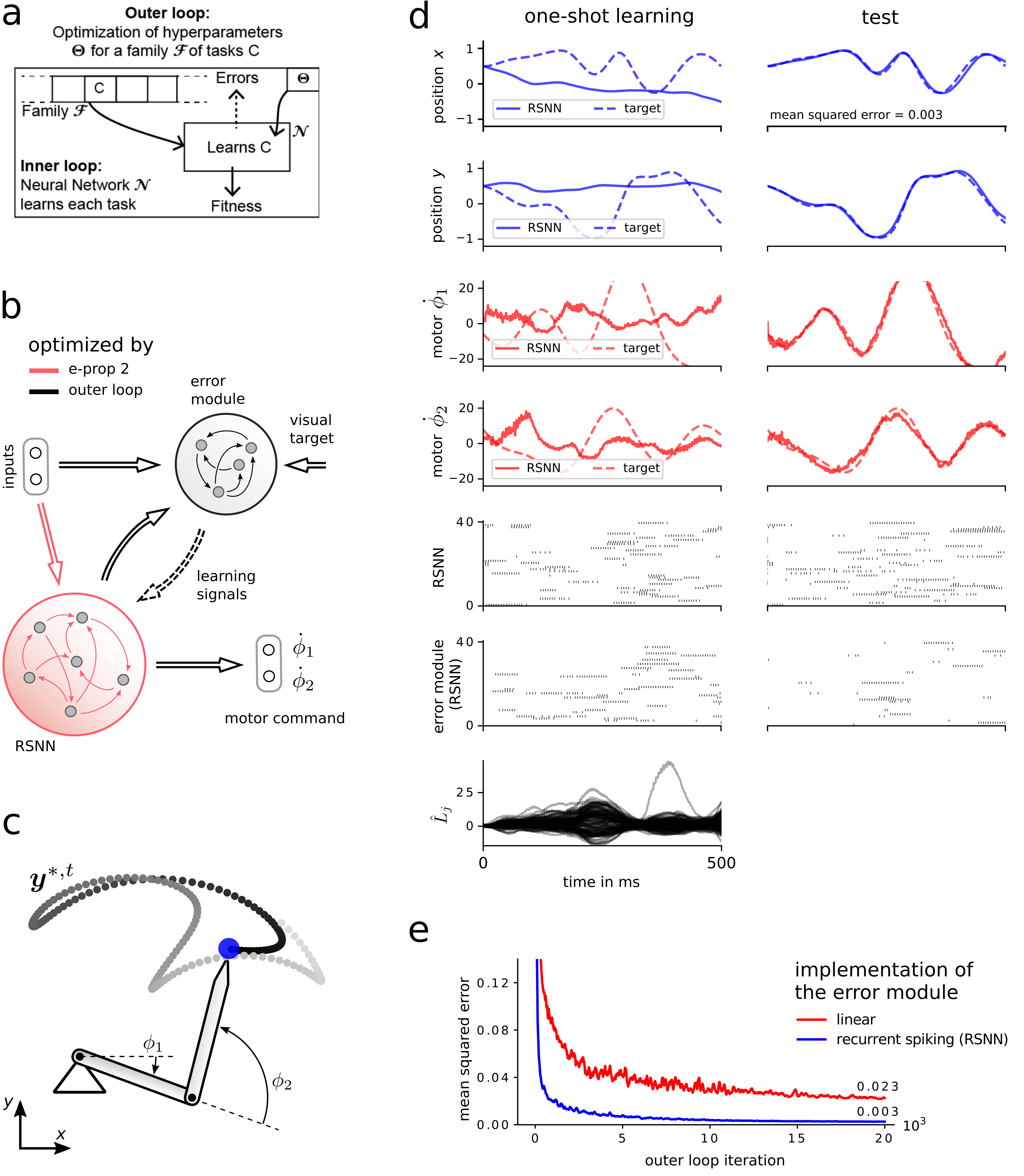}
	\caption{\textbf{Scheme and performance of \textit{e-prop 2}} 
	\textbf{a})~Learning-to-Learn (LTL) scheme. 
	\textbf{b})~Learning architecture for \textit{e-prop 2}. In this demo the angular velocities of the joints were controlled by a recurrent network of spiking neurons (RSNN). 
	A separate error module was optimized in the outer loop of L2L to produce suitable learning signals. \textbf{c})~Randomly generated target movements $\bm y^{*,t}$ (example shown) had to be reproduced by the tip of an arm with two joints. \textbf{d})~Demonstration of one-shot learning for a randomly sampled target movement. During the training trial the error module sends learning signals (bottom row) to the network. After a single weight update the target movement can be reproduced in a test trial with high precision. \textbf{e})~One-shot learning performance improved during the course of outer loop optimization. Two error module implementations were compared.}
	\label{fig:one-shot}
\end{figure}

\subsection*{\textit{E-prop 2}: Refined learning signals that emerge from L2L}
\label{sec:e-prop2}
The construction and proper distribution of learning signals appears to be highly sophisticated in the brain. Numerous areas in the human brain appear to be involved in the production of the error-related negativity and the emission of neuromodulatory signals (see the references given in the Introduction). A closer experimental analysis suggests diversity and target-specificity even for a single neuromodulator \citep{engelhard2018specialized}. Hence it is fair to assume that the construction and distribution of error signals in the brain has been optimized through evolutionary processes, through development, and prior learning. A simple approach for capturing possible consequences of such an optimization in a model is to apply L2L to a suitable family of learning tasks. The outer loop of L2L models opaque optimization processes that shape the distribution of error signals in the brain on the functional level. We implement this optimization by an application of BPTT to a separate error module in the outer loop of L2L. Since this outer loop is not meant to model an online learning process, we are not concerned here by the backpropagation through time that is required in the outer loop. In fact, one can expect that similar results can be achieved through an application of gradient-free optimization methods in the outer loop, but at a higher computational cost for the implementation.

It is argued in \citep{BreaGerstner:16} that one-shot learning is one of two really important learning capabilities of the brain that are not yet satisfactorily explained by current models in computational neuroscience. We show here that \textit{e-prop 2} explains how RSNNs can learn a new movement trajectory in a single trial. Simultaneously we show that given movement trajectories of an end-effector of an arm model can be learnt without requiring an explicitly learnt or constructed inverse model. Instead, a suitably trained error module can acquire the capability to produce learning signals for the RSNN so that the RSNN learns via \textit{e-prop} to minimize deviations of the end-effector from the target trajectory in Euclidean space, rather than errors in ``muscle space", i.e., in terms of the joint angles that are controlled by the RSNN.

\textbf{Definition of \textit{e-prop 2}:}
The main characteristic of \textit{e-prop 2} is that the learning signals $\widehat{L}_j^t$ are produced by a trained error module, which is modeled as a recurrent network of spiking neurons with synaptic weights $\bm \Psi$. It receives the input $\bm x^t$, the spiking activity in the network $\bm z^t$ and target signals $\bm y^{*,t}$. Note that the target signal is not necessarily the target output of the network, but can be more generally a target state vector of some controlled system. 

We employ the concept of Learning-to-Learn (L2L) to enable the network with its adjacent error module to solve a family $\mathcal{F}$ of one-shot learning tasks, i.e. each task $C$ of the family $\mathcal{F}$ requires the network to learn a movement from a single demonstration. The L2L setup introduces a nested optimization procedure that consists of two loops: An inner loop and an outer loop as illustrated in Figure~\ref{fig:one-shot}a. In the inner loop we consider a particular task $C$, entailing a training trial and a testing trial. During the training trial, the network has synaptic weights $\bm \theta_\mathrm{init}$, it receives an input it has never seen and generates a tentative output. After a single weight update using \textit{e-prop 2}, the network starts the testing trial with new weights $\bm \theta_{\mathrm{test},C}$. It receives the same input for a second time and its performance is evaluated using the cost function $\mathcal{L}_C( \bm \theta_{\mathrm{test},C})$. In the outer loop we optimize $\bm \theta_\mathrm{init}$ and the error module parameters $\bm \Psi$ in order to minimize the cost $\mathcal{L}_C$ over many task instances $C$ from the family $\mathcal{F}$. Formally, the optimization problem solved by the outer loop is written as:

\begin{eqnarray}
\min_{\bm \Psi, \bm \theta_\mathrm{init}} & \mathds{E}_{C \sim \mathcal{F}}\left[ \mathcal{L}_C( \bm \theta_{\mathrm{test},C}) \right] \\
\text{s.t.:} & \left( \bm \theta_{\mathrm{test},C} \right)_{ji} = \left(\bm \theta_\mathrm{init} \right)_{ji} -\eta \sum_t  \hat L_j^t ~ e_{ji}^t \label{eq:ltl_weight_update}
\\ & \text{($ \hat L_j^t$ and $e_{ji}^t$ are obtained during the training} \nonumber \\ & \text{ trial for task $C$ using $\bm \Psi$ and $\theta_\mathrm{init}$)}, \nonumber
\end{eqnarray}
where $\eta$ represents a fixed learning rate.

\paragraph{One-shot learning task 2.1}
It is likely that prior optimization processes on much slower time scales, such as evolution and development, have prepared many species of animals to learn new motor skills much faster than shown in Figure~\ref{fig:partials}d. 
Humans and other species can learn a new movement by observing just one or a few examples. 
Therefore 
we restricted the learning process here to a single trial (one-shot learning, or imitation learning).

Another challenge for motor learning arises from the fact that motor commands $\dot \Phi^t$ have to be given in ``muscle space" (joint angle movements), whereas
the observed resulting movement $\bm y^t$ is given in Euclidean space. 
Hence an inverse model is usually assumed to be needed to infer joint angle movements that can reduce the observed error. We show here that an explicit inverse model is not needed, since its function can be integrated into the learning signals from the error module of \textit{e-prop 2}.

\textbf{Task:}
Each task $C$ in the family $\mathcal{F}$ consisted of learning a randomly generated target movement $\bm y^{*,t}$ of the tip of a two joint arm as shown in Figure~\ref{fig:one-shot}c. 
The task was divided into a training and a testing trial, with a single weight update in between according to equation~\eqref{eq:ltl_weight_update}.

\textbf{Implementation:}
An RSNN, consisting of 400 recurrently connected LIF neurons, learnt to generate the required motor commands, represented as the angular velocities of the joints $\dot \Phi^t = (\dot \phi_1^t, \dot \phi_2^t)$, in order to produce the target movement. The full architecture of the learning system is displayed in Figure~\ref{fig:one-shot}b.
The error module consisted of 300 LIF neurons, which were also recurrently connected.
The input $\bm x^t$ to the network  was the same across all trials and was given by a clock-like signal. 
The 
input to the error module contained a copy of 
$\bm x^t$, the spiking activity $\bm z^t$ of the main network, as well as the target movement $\bm y^{*,t}$ in Euclidean space. Importantly, the error module had no access to actual errors of the produced motor commands.
For outer loop optimization we viewed the learning process as a dynamical system for which we applied BPTT. Gradients were computed using batches of different tasks to approximate the expectation in the outer loop objective.

\textbf{Performance:}
After sufficiently long training in the outer loop of L2L, we tested the learning capabilities of \textit{e-prop 2} on a random target movement, and show in Figure~\ref{fig:one-shot}d training and testing trial in the left and right column respectively. 
In fact, after the error module had sent learning signals to the network during the training trial, it was usually more silent during testing, since the reproduced movement was already accurate.
Therefore, the network was endowed with one-shot learning capabilities by \textit{e-prop 2}, after initial weights $\bm \theta_\mathrm{init}$ and the error module parameters $\bm \Psi$ had been optimized in the outer loop. 

Figure~\ref{fig:one-shot}e summarizes the mean squared error between the target $\bm y^{*,t}$ and actual movement $\bm y^t$ in the testing trial (blue curve). The red curve reports the same for a linear error module. The error is reported for different stages of the outer loop optimization. 

We considered also the case when one uses instead of the eligibility trace as defined in equation~\eqref{eq:elig-scalar-LIF} just a truncated one, given by $e_{ji}^t = h_j^t z_i^{t-1}$, and found this variation to exhibit similar performance on this task (not shown). The learning performance converged to a mean squared error of $0.005$ on testing trials averaged over different tasks.


Altogether we have shown that \textit{e-prop 2} enables one-shot learning of pattern generation by an RSNN. This is apparently the first time that one-shot learning of pattern generation has been demonstrated for an RSNN. In addition, we have shown that the learning architecture for \textit{e-prop 2} supports a novel solution to motor learning, where no separate construction or learning of an inverse model is required. More precisely, the error module can be trained to produce learning signals that enable motor learning without the presence of an inverse model. It will be interesting to see whether there are biological data that support this simplified architecture. Another interesting feature of the resulting paradigm for motor learning is that the considered tasks can be accomplished without sensory feedback about the actual trajectory of the arm movement. Instead the error module just receives efferent copies of the spiking activity of the main network, and hence implicitly also of its motor commands.

\begin{figure}
	\centering
	\includegraphics[width=\textwidth]{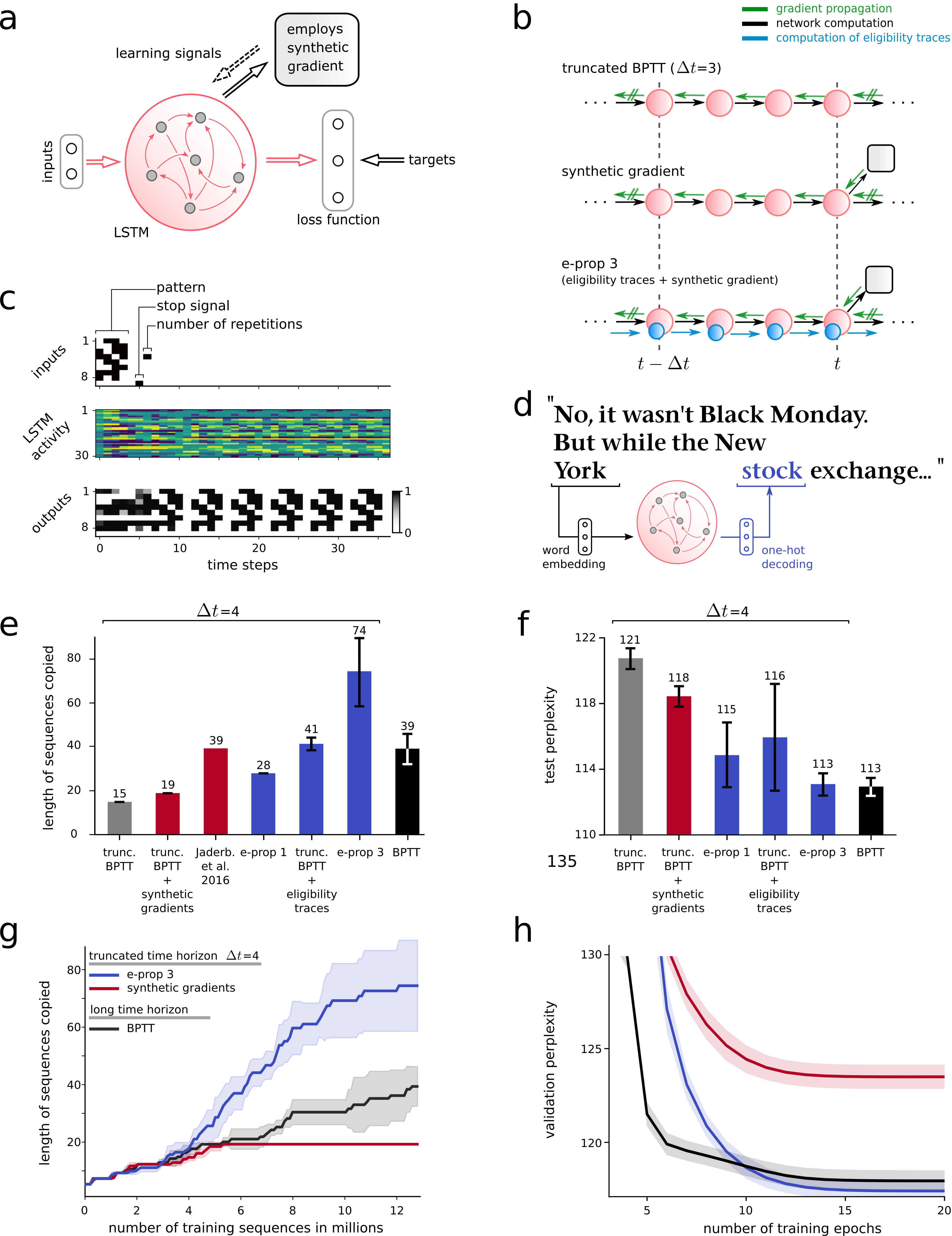}
	\caption{\textbf{Scheme and performance of \textit{e-prop 3}} 
		\textbf{a})~Learning architecture for \textit{e-prop 3}. The error module is implemented through synthetic gradients.
		\textbf{b})~Scheme of learning rules used in panels \textbf{e-h}. $\Delta t$ is the number of time steps through which the gradients are allowed to flow for truncated BPTT or for the computation of synthetic gradients.
		\textbf{c},\textbf{e},\textbf{g}) Copy-repeat task: 
		(\textbf{c}) example trial, 
		(\textbf{e}) performance of different algorithms for the copy-repeat task, 
		(\textbf{g}) Learning progress for 3 different algorithms.
		This task was used as a benchmark in \citep{jaderberg2016decoupled} and \citep{graves2014neural}.
		\textbf{d},\textbf{f},\textbf{h}) Word prediction task: (\textbf{d}) example of sequence, (\textbf{f}) performance summary for different learning rules, (\textbf{h}) learning progression.
		An epoch denotes a single pass through the complete dataset.
	}
	\label{fig:copy}
\end{figure}

\subsection*{\textit{E-prop 3}: Producing learning signals through synthetic gradients}
We show here that the use of biologically inspired eligibility traces also improves some state-of-the-art algorithms in machine learning, specifically the synthetic gradient approach for learning in feedforward and recurrent (artificial) neural networks \citep{jaderberg2016decoupled} and \citep{czarnecki2017understanding}. Synthetic gradients provide variants of backprop and BPTT that tend to run more efficiently because they avoid that synaptic weights can only be updated after a full network simulation followed by a full backpropagation of error gradients („locking“). We show here that the performance of synthetic gradient methods for recurrent neural networks significantly increases when they are combined with eligibility traces. The combination of these two approaches is in a sense quite natural, since synthetic gradients can be seen as online learning signals for \textit{e-prop}. In comparison with \textit{e-prop 2}, one does not need to consider here a whole family of learning tasks for L2L and a computationally intensive outer loop. Hence the production of learning signals via synthetic gradient approaches tends to be computationally more efficient. So far it also yields better results in applications to difficult tasks for recurrent artificial neural networks. In principle it is conceivable that biological learning systems also follow a strategy whereby learning signals for different temporal phases of a learning process are aligned among each other through some separate process. This is the idea of synthetic gradients.

\textbf{Definition of \textit{e-prop 3}:}
When BPTT is used for tasks that involve long time series, the algorithm is often truncated to shorter intervals of length $\Delta t$, and the parameters are updated after the processing of each interval.
This variant of BPTT is called truncated BPTT. We show a schematic of the computation performed on the interval $\{t - \Delta t,\dots, t\}$ in the first panel of Figure \ref{fig:copy}b.
Reducing the length $\Delta t$ of the interval has two benefits: firstly, the parameter updates are more frequent; and secondly, it requires less memory storage because all inputs $\bt x^{t'}$ and network states $\bt s_j^{t'}$ for $t' \in \{t - \Delta t,\dots, t\}$ need to be stored temporarily for back-propagating gradients. The drawback is that the error gradients become more approximative, in particular, the relationship between error and network computation happening outside of this interval cannot be captured by the truncated error gradients. As illustrated in the last panel of Figure \ref{fig:copy}b, \textit{e-prop 3} uses the same truncation scheme but alleviates the drawback of truncated BPTT in two ways: it uses eligibility traces to include information about the network history before $t-\Delta t$, and an error module that predicts errors after $t$.
Thanks to eligibility traces, all input and recurrent activity patterns that happened before $t-\Delta t$ have left a footprint that can be combined with the learning signals $L_j^{t'}$ with $t' \in \{t - \Delta t,\dots, t\}$. Each of these learning signals summarizes the neuron influence on the next errors, but due to the truncation, errors performed outside of the current interval cannot trivially be taken into account. In \textit{e-prop 3}, the error module computes learning signals $L_j^{t'}$ that anticipate the predictable components of the future errors.

\textit{E-prop 3} combines eligibility traces and learning signals to compute error gradients according to equation \eqref{eq:grad}, rather than equation \eqref{eq:bptt} that is canonically used to compute gradients for BPTT. To compute these gradients when processing the interval $\{t - \Delta t,\dots, t\}$, the eligibility traces and learning signals are computed solely from data available within that interval, without requiring the rest of the data. The eligibility traces are computed in the forward direction according to equations \eqref{eq:elig-vector} and \eqref{eq:elig-scalar}.
For the learning signals $L_j^{t'}$, the difficulty is to estimate the gradients $\frac{dE}{dz_j^{t'}}$ for $t'$ between $t-\Delta t$ and $t$. These gradients are computed by back-propagation from $t'=t$ back to $t'=t-\Delta t +1$ with the two recursive formulas:
\begin{eqnarray}
  \frac{d E}{d \bt s_j^{t'}}
	& = & \frac{d E}{d z_j^{t'}} \frac{\partial z_j^{t'}}{\partial \bt s_j^{t'}} + \frac{d E}{d \bt s_j^{t'+1}} \frac{\partial \bt s_j^{t'+1}}{\partial \bt s_j^{t'}} \label{eq:backpropagation-results} \\
  \frac{d E}{d z_j^{t'}} & = & \frac{\partial E}{\partial z_j^{t'}} + \sum_i \frac{d E}{d \bt s_i^{t'+1}} \frac{\partial \bt s_i^{t'+1}}{\partial z_j^{t'}}~,
\end{eqnarray}
which are derived by application of the chain rule at the nodes $\bt s_j^t$ and $z_j^t$ of the computational graph represented in Figures \ref{fig:graph} ($\frac{\partial \bt s_i^{t+1}}{\partial z_j^{t}}$ is a notation short-cut for $\frac{\partial M}{\partial z_j^{t}}(\bt s_i^t, \bt z^t , \bt x^t , \bth)$, and $i$ range over all neurons to which neuron $j$ is synaptically connected).
This leaves open the choice of the boundary condition $\frac{d E}{d \bt s_j^{t+1}}$ that initiates the back-propagation of these gradients at the end of the interval $\{t - \Delta t,\dots, t\}$.
In most implementations of truncated BPTT, one chooses $\frac{d E}{d \bt s_j^{t+1}}=0$, as if the simulation would terminate after time $t$.
We use instead a feed-forward neural network $\SG$ parametrized by $\Psi$ that outputs a boundary condition $\SG_j$ associated with each neuron $j$: $\frac{d E}{d \bt s_j^{t+1}}= \SG_j(\bt z^t, \Psi)$. This strategy was proposed by \citep{jaderberg2016decoupled} under the name ``synthetic gradients''.
The synthetic gradients are combined with the error gradients computed within the interval $\{t - \Delta t,\dots, t\}$ to define the learning signals.
Hence, we also see $\SG$ as a key component of the an error module in analogy with those of \textit{e-prop 1} and 2.

To train $\SG$, back-ups of the gradients $\frac{dE}{d \bt s_j^{t+1}}$ are estimated from the boundary conditions at the end of the next interval $\{t,\dots,t+\Delta t\}$. In a similar way as value functions are approximated in reinforcement learning, these more informed gradient estimates are used as targets to improve the synthetic gradients $\SG(\bt z^t, \Psi)$. This is done in \textit{e-prop 3} by computing simultaneously the gradients $\frac{d E'}{d\bth}$ and $\frac{d E'}{d\Psi}$ of an error function $E'$ that combines the error function $E$, the boundary condition, and the mean squared error between the synthetic gradient and its targeted back-up (see Algorithm \ref{alg:sg} in Methods for details). These gradients are approximated to update the network parameters with any variant of stochastic gradient descent.

It was already discussed that the factorization \eqref{eq:grad} used in \textit{e-prop} is equivalent to BPTT, and that both compute the error gradients $\frac{dE}{d\theta_{ji}}$. In the subsection dedicated to \textit{e-prop 3} of Methods, we formalize the algorithm when the simulation is truncated into intervals of length $\Delta t$.
We then show under the assumption that the synthetic gradients are optimal, i.e. $\SG_j(\bt z^t, \Psi)= \frac{d E}{d {\bt s_j^{t+1}}}$, that both truncated BPTT and \textit{e-prop 3} compute the correct error gradients $\frac{dE}{d \theta_{ji}}$.
However, this assumption is rarely satisfied in simulations, firstly because the parameters $\Psi$ may not converge instantaneously to some optimum; and even then, there could be unpredictable components of the future errors that cannot be estimated correctly. Hence, a more accurate model is to assume that the synthetic gradients $\SG_j(\bt z^t, \Psi)$ are noisy estimators of $\frac{d E}{d {\bt s_j^{t+1}}}$. Under this weaker assumption it turns out that \textit{e-prop 3} produces estimators of the error gradients $\widehat{\frac{dE}{d\theta_{ji}}}^{\mathrm{eprop}}$ that are better than those produced with truncated BPTT with synthetic gradients $\widehat{\frac{dE}{d\theta_{ji}}}^{\mathrm{SG}}$. Formally, this result can be summarized as:
\begin{equation}
\EE \left[ \left( \frac{dE}{d\theta_{ji}} - \widehat{\frac{dE}{d\theta_{ji}}}^{\mathrm{eprop}} \right) ^2 \right] \leq 
\EE \left[ \left( \frac{dE}{d\theta_{ji}} - \widehat{\frac{dE}{d\theta_{ji}}}^{\mathrm{SG}} \right) ^2 \right] ~,
\end{equation}
where the $\EE$ is the stochastic expectation.
The proof of this result will be published in a later version of the paper.
To summarize the proof, we compare in detail the terms computed with \textit{e-prop 3} and BPTT. The derivation reveals that both algorithms compute a common term that combines partial derivatives $\frac{\partial \bt s^{t}}{\partial \theta_{ji}}$ with errors, $E(\bt z^{t'})$ with $t$ and $t'$ being accessible within the interval $\{t - \Delta t,\dots, t\}$. The difference between the two algorithms is that truncated BPTT combines all the partial derivatives $\frac{\partial \bt s^{t}}{\partial \theta_{ji}}$ with synthetic gradients that predict future errors. Instead, the \textit{e-prop} algorithm holds these partial derivatives in eligibility traces to combine them later with a better informed learning signals that do not suffer from the noisy approximations of the synthetic gradients.

\paragraph{Copy-repeat task 3.1}
The copy-repeat task was introduced in \citep{graves2014neural} to measure how well artificial neural networks can learn to memorize and process complex patterns. It was also used in \citep{jaderberg2016decoupled} to compare learning algorithms for recurrent neural networks with truncated error propagation.

\textbf{Task:} The task is illustrated in Figure \ref{fig:copy}c. It requires to read a sequence of 8-bit characters followed by a ``stop'' character and a character that encodes the requested number of repetitions. In the subsequent time steps the network is trained to repeat the given pattern as many times as requested, followed by a final ``stop'' character. We used the same curriculum of increased task complexity as defined in \citep{jaderberg2016decoupled}, where the authors benchmarked variants of truncated BPTT and synthetic gradients: the pattern length and the number of repetitions increase alternatively each time the network solves the task (the task is considered solved when the average error is below $0.15$ bits per sequence). The performance of the learning algorithms are therefore measured by the length of the largest sequence for which the network could solve the task.

\textbf{Implementation:} The recurrent network consisted of 256 LSTM units. The component $\SG$ of the error module was a feedfoward network with one hidden layer of 512 rectified linear units (the output layer of the synthetic gradient did not have non-linear activation functions).
The network states were reset to zero at the beginning of each sequence and the mean error and the error gradients were averaged over a batch of 256 independent sequences. Importantly, we used for all training algorithms a fixed truncation length $\Delta t=4$, with the exception of a strong baseline BPTT for which the gradients were not truncated.

\textbf{Results:}
The activity and output of a trained LSTM solving the task is displayed in Figure \ref{fig:copy}c. The performance of various learning algorithms is summarized in Figure \ref{fig:copy}e and g.
Truncated BPTT alone solves the task for sequences of $15$ characters, and $19$ characters when enhanced with synthetic gradients. With a different implementation of the task and algorithm, \citep{jaderberg2016decoupled} reported that sequences of $39$ characters could be handled with synthetic gradients.
When BPTT is replaced by \textit{e-prop} and uses eligibility traces, the network learnt to solve the task for sequences of length $41$ when the synthetic gradients were set to zero (this algorithm is referred as ``truncated BPTT + eligibility traces'' in Figure \ref{fig:copy}e and f). The full \textit{e-prop 3} algorithm that includes eligibility traces and synthetic gradients solved the task for sequences of $74$ characters.
This is an improvement over \textit{e-prop 1} that handles sequences of $28$ characters, even if the readout weights are not replaced by random error broadcasts.
All these results were achieved with a truncation length $\Delta t=4$. In contrast when applying full back-propagation through the whole sequence, we reached only $39$ characters.

\paragraph{Word prediction task 3.2}
We also considered a word-level prediction task in a corpus of articles extracted from the Wall Street Journal, provided by the so-called Penn Treebank dataset. As opposed to the previous copy-repeat task, this is a practically relevant task. It has been standardized to provide a reproducible benchmark task.
Here, we used the same implementation and baseline performance as provided freely by the Tensorflow tutorial on recurrent neural networks\footnote{https://www.tensorflow.org/tutorials/sequences/recurrent}.

\textbf{Task:}
As indicated in Figure \ref{fig:copy}d, the network reads the whole corpus word after word. At each time step, the network has to read a word and predict the following one.
The training, validation and test sets consist of texts of 929k, 73k, and 82k words.
The sentences are kept in a logical order such that the context of dozens of words matters to accurately predict the following ones.
The vocabulary of the dataset is restricted to the 10k most frequent words, and the words outside of this vocabulary are replaced with a special unknown word token.

\textbf{Implementation:}
For all algorithms, the parameters were kept identical to those defined in the Tensorflow tutorial, with two exceptions: firstly, the networks had a single layer of 200 LSTM units instead of two to simplify the implementation, and because the second did not seem to improve performance with this configuration; secondly, the truncation length was reduced from $\Delta t=20$ to $\Delta t=4$ for synthetic gradients and \textit{e-prop 3} to measure how our algorithms compensate for it.
The synthetic gradients are computed by a one hidden layer of 400 rectified linear units.
For a detailed description of model and a list of parameters we refer to the methods.

\textbf{Results:}
The error is measured for this task by the word-level perplexity, which is the exponential of the mean cross-entropy loss.
Figure \ref{fig:copy}e and f summarize our results.
After reduction to a single layer, the baseline perplexity of the model provided in the Tensorflow tutorial for BPTT was $113$ with a truncation length $\Delta t=20$.
Full BPTT is not practical in this case because the data consists of one extremely long sequence of words.
In contrast, in a perplexity increased to $121$ when the same model was also trained with BPTT, but a shorter truncation length $\Delta t=4$.
The model performance improved back to $118$ with synthetic gradients, and to $116$ with eligibility traces.
Applying the \textit{e-prop 1} algorithm with the true readout weights as error broadcasts resulted in a perplexity of $115$.
When combining both eligibility traces and synthetic gradients in \textit{e-prop 3}, the performance improved further and we achieved a perplexity of $113$.

To further investigate the relevance of eligibility traces for \textit{e-prop 3} we considered the case where the eligibility trace were truncated.
Instead of using the eligibility trace vectors as defined in equation~\eqref{eq:elig-vector-lstm} we used $\bt \bepsi_{ji}^t = \frac{\partial \bt s_j^t}{\partial \theta_{ji}}$. This variation of \textit{e-prop 3} lead to significantly degraded performance and resulted in test perplexity of $122.67$ (not shown).

All together Figure \ref{fig:copy}e and f show that eligibility traces improve truncated BPTT more than synthetic gradients. Furthermore, if eligibility traces are combined with synthetic gradients in \textit{e-prop 3}, one arrives at an algorithm that outperforms full BPTT for the copy-repeat task, and matches the performance of BPTT ($\Delta t=20$) for the Penn Treebank word prediction task.

\section*{Discussion}

The functionally most powerful learning method for recurrent neural nets, an approximation of gradient descent for a loss function via BPTT, requires propagation of error signals backwards in time. Hence this method does not reveal how recurrent networks of neurons in the brain learn. In addition, propagation of error signals backwards in time requires costly work-arounds in software implementations, and it does not provide an attractive blueprint for the design of learning algorithms in neuromorphic hardware. We have shown that a replacement of the propagation of error signals backwards in time in favor of a propagation of the local activation histories of synapses -- called eligibility traces -- forward in time allows us to capture with physically and biologically realistic mechanisms a large portion of the functional benefits of BPTT. We are referring to to this new approach to gradient descent learning in recurrent neural networks as \textit{e-prop}. In contrast to many other approaches for learning in recurrent networks of spiking neurons it can be based on a rigorous mathematical theory.

We have presented a few variations of \textit{e-prop} where eligibility traces are combined with different types of top-down learning signals that are generated and transmitted in real-time. In \textit{e-prop 1} we combine eligibility traces with a variation of broadcast alignment \citep{samadi2017deep} or direct feedback alignment \citep{nokland2016direct}.
We first evaluated the performance of \textit{e-prop 1} on a task that has become a standard for the evaluation of the FORCE learning method for recurrent networks of spiking neurons \citep{nicola2017supervised} and related earlier work on artificial neural networks: supervised learning of generating a temporal pattern. In order to make the task more interesting we considered a task where 3 independent temporal patterns have to be generated simultaneously by the same RSNN. Here it is not enough to transmit a single error variable to the network, so that broadcasting of errors for different dimensions of the network output to the network becomes less trivial. We found (see Figure~\ref{fig:partials}) that a random weight matrix for the distribution of error signals works well, as in the case of feedforward networks \citep{samadi2017deep}, \citep{nokland2016direct}. But surprisingly, the results were best when this matrix was fixed, or rarely changed, whereas a direct application of broadcast alignment to an unrolled recurrent network suggests that a different random matrix should be used for every time slice.

In order to challenge the capability of \textit{e-prop 1} to deal also with cases where error signals arise only at the very end of a computation in a recurrent network, we considered a store-recall task, where the network has to learn what information it should store --and maintain until it is needed later. We found (see Figure~\ref{fig:store_recall}) that this task can also be learnt with \textit{e-prop 1}, and verified that the eligibility traces of the network were able to bridge the delay. We used here an LSNN \citep{bellec2018long} that includes a model of a slower process in biological neurons: neuronal adaptation. We also compared the performance of \textit{e-prop 1} with BPTT for a more demanding task: the speech recognition benchmark task TIMIT. We found that \textit{e-prop 1} approximates also here the performance of BPTT quite well.

Altogether we have the impression that \textit{e-prop 1} can solve all learning tasks for RSNNs that the FORCE method can solve, and many more demanding tasks. Since the FORCE method is not argued to be biologically realistic, whereas \textit{e-prop 1} only relies on biologically realistic mechanisms, this throws new light on the understanding of learning in recurrent networks of neurons in the brain. An additional new twist is that \textit{e-prop 1} engages also plasticity of synaptic connections within a recurrent network, rather than only synaptic connections to a postulated readout neuron as in the FORCE method. Hence we can now analyze how network configurations and motifs that emerge in recurrent neural network models through learning (possibly including in e-prop biologically inspired rewiring mechanisms as in \citep{bellec2018deep}) relate to experimental data.
  
In the analysis of \textit{e-prop 2} we moved to a minimal model that captures salient aspects of the organization of learning in the brain, where dedicated brain areas process error signals and generate suitably modified gating signals for plasticity in different populations of neurons. We considered in this minimal model just a single RSNN (termed error module) for generating learning signals. But obviously this minimal model opens the door to the analysis of more complex learning architectures --as they are found in the brain-- from a functional perspective. We found that a straightforward application of the Learning-to-Learn (L2L) paradigm, where the error module is optimized on a slower time scale for its task, significantly boosts the learning capability of an RSNN. Concretely, we found that it endows the RSNN with one-shot learning capability (see Figure \ref{fig:one-shot}), hence with a characteristic advantage of learning in the human brain \citep{BreaGerstner:16}, \citep{lake2017building}. In addition the model of Figure \ref{fig:one-shot} suggests a new way of thinking about motor learning. It is shown that no separate inverse model is needed to learn motor control. Furthermore in the case that we considered, not even sensory feedback from the environment is needed.
 
Finally we arrived at an example where biologically inspired ideas and mechanisms, in this case eligibility traces, can enhance state-of-the-art methods in machine learning. Concretely, we have shown in Figure \ref{fig:copy} that adding eligibility traces to the synthetic gradient methods of \citep{jaderberg2016decoupled} and \citep{czarnecki2017understanding} for training artificial recurrent neural networks significantly enhances the performance of synthetic gradient algorithms. In fact, the resulting algorithm \textit{e-prop 3} was found to supercede the performance of full BPTT in one case and rival BPTT with $\Delta t=20$ in another.

A remarkable feature of \textit{e-prop} is that the resulting local learning rules \eqref{eq:grad-LIF} are very similar to previously proposed rules for synaptic plasticity that were fitted to experimental data \citep{clopath2010connectivity}.
In fact, we have shown in Figure \ref{fig:partials}d that the theory-derived local plasticity rule for \textit{e-prop 1} can be replaced by the Clopath rule with little loss in performance. On a more general level, the importance of eligibility traces for network learning that our results suggest provides concrete hypotheses for the functional role of a multitude of processes on the molecular level in neurons and synapses, including metabotropic receptors. Many of these processes are known to store information about multiple aspects of the recent history. The \textit{e-prop} approach suggests that these processes, in combination with a sufficiently sophisticated production of learning signals by dedicated brain structures, can practically replace the physically impossible backpropagation of error signals backwards in time of theoretically optimal BPTT.

An essential prediction of e-prop for synaptic plasticity rules is that learning signals can switch the sign of synaptic plasticity, i.e., between LTP and LTD or between STDP and anti-STDP. Such switching of the sign of plasticity via disinhibition had been found in synapses from the cortex to the striatum \citep{paille2013gabaergic}, see \citep{perrin2019bridging} for a recent review. Further brain mechanisms for switching the sign of plasticity through 3rd factors had been reported in \citep{chen2014short, cui2016endocannabinoid, foncelle2018modulation}.

A key challenge for neuromorphic engineering is to design a new generation of computing hardware that enables energy efficient implementations of major types of networks and learning algorithms that have driven recent progress in machine learning and learning-driven AI (see e.g. \citep{barrett2018measuring}). Recurrent neural networks are an essential component of many of these networks, and hence learning algorithms are needed for this type of networks that can be efficiently implemented in neuromorphic hardware. In addition, neuromorphic implementations of recurrent neural networks -- rather than deep feedforward networks—promise larger efficiency gains because hardware neurons can be re-used during a computation. Recently developed diffusive memristors \citep{wang2018fully} would facilitate an efficient local computation of eligibility traces with new materials. In addition, new 3-terminal memristive synapses \citep{yang2017multifunctional} are likely to support an efficient combination of local eligibility traces with top-down error signals in the hardware. Thus \textit{e-prop} provides attractive functional goals for novel materials in neuromorphic hardware.

\section*{Methods}

\paragraph{General network model:}
Our proposed learning algorithms for recurrent neural networks can be applied to a large class of spiking and non-spiking neural network models. 
We assume that the state at time $t$ of each neuron $j$ in the network can be described by an internal state vector $\bt s_j^t \in \RR^d$ and an observable state $z_j^t$. The internal state includes internal variables of the neuron such as its activation or membrane potential. The observable state is given by the output of the neuron (analog output for ANNs and spiking output for SNNs). The dynamics of a neuron's discrete-time state evolution is described by two functions $M(\bt s, \bt z, \bt x, \boldsymbol \theta)$ and $f(\bt s)$, where $\bt s$ is an internal state vector, $\bt z$ is a vector of the observable network state (i.e., outputs of all neurons in the network), $\bt x$ is the vector of inputs to the network, and $\boldsymbol \theta$ denotes the vector of network parameters (i.e., synaptic weights). In particular, for each neuron $j$ the function $M$ maps from the current network state observable to that neuron to its next internal state, and $f$ maps from its internal state to its observable state (neuron output):
\begin{eqnarray}
  \bt s_j^{t} &=& M(\bt s_j^{t-1}, \bt z^{t-1}, \bt x^t, \boldsymbol \theta),\\
  z_j^t &=& f(\bt s_j^t),
\end{eqnarray}
where $\bt z^t$ ($\bt x^t$) denotes the vector of observable states of all network (input) neurons at time $t$. A third function $E$ defines the error of the network within some time interval $0,\dots, T$. It is assumed to depend only on the observable states $E(\bt z^1, \dots, \bt z^T)$. 

We explicitly distinguish betweens partial derivative and total derivatives in our notation. We write $\frac{\partial M}{\partial s}(\bt s^*, \bt z^*, \bt x^*, \boldsymbol \theta)$ to denote the partial derivative of the function $M$ with respect to $\bt s$, applied to particular arguments $\bt s^*, \bt z^*, \bt x^*, \boldsymbol \theta$.
To simplify notation, we define the shortcuts $\frac{\partial \bt s_j^{t}}{\partial \bt s_j^{t-1}} \eqdef \frac{\partial M}{\partial s}(\bt s_j^{t-1}, \bt z^{t-1}, \bt x^t, \boldsymbol \theta)$, 
$\frac{\partial \bt s_j^{t}}{\partial \theta_{ji}} \eqdef \frac{\partial M}{\partial \theta_{ji}}(\bt s_j^{t-1}, \bt z^{t-1}, \bt x^t, \boldsymbol \theta)$, and
$\frac{\partial z_j^t}{\partial \bt s_j^{t}} \eqdef \frac{\partial f}{\partial s}(\bt s_j^{t})$.

To emphasize that $\frac{\partial \bt s_j^{t}}{\partial \bt s_j^{t-1}}$ is a matrix of shape $d \times d$, and because it has an important role in the following derivation and in definition of eligibility traces, we also use the further notation $D_j^t=\frac{\partial \bt s_j^{t+1}}{\partial \bt s_j^{t}}$.
Note that we write gradients as row vectors and states as column vectors.

\begin{figure}
    \centering
    \includegraphics[width=\textwidth]{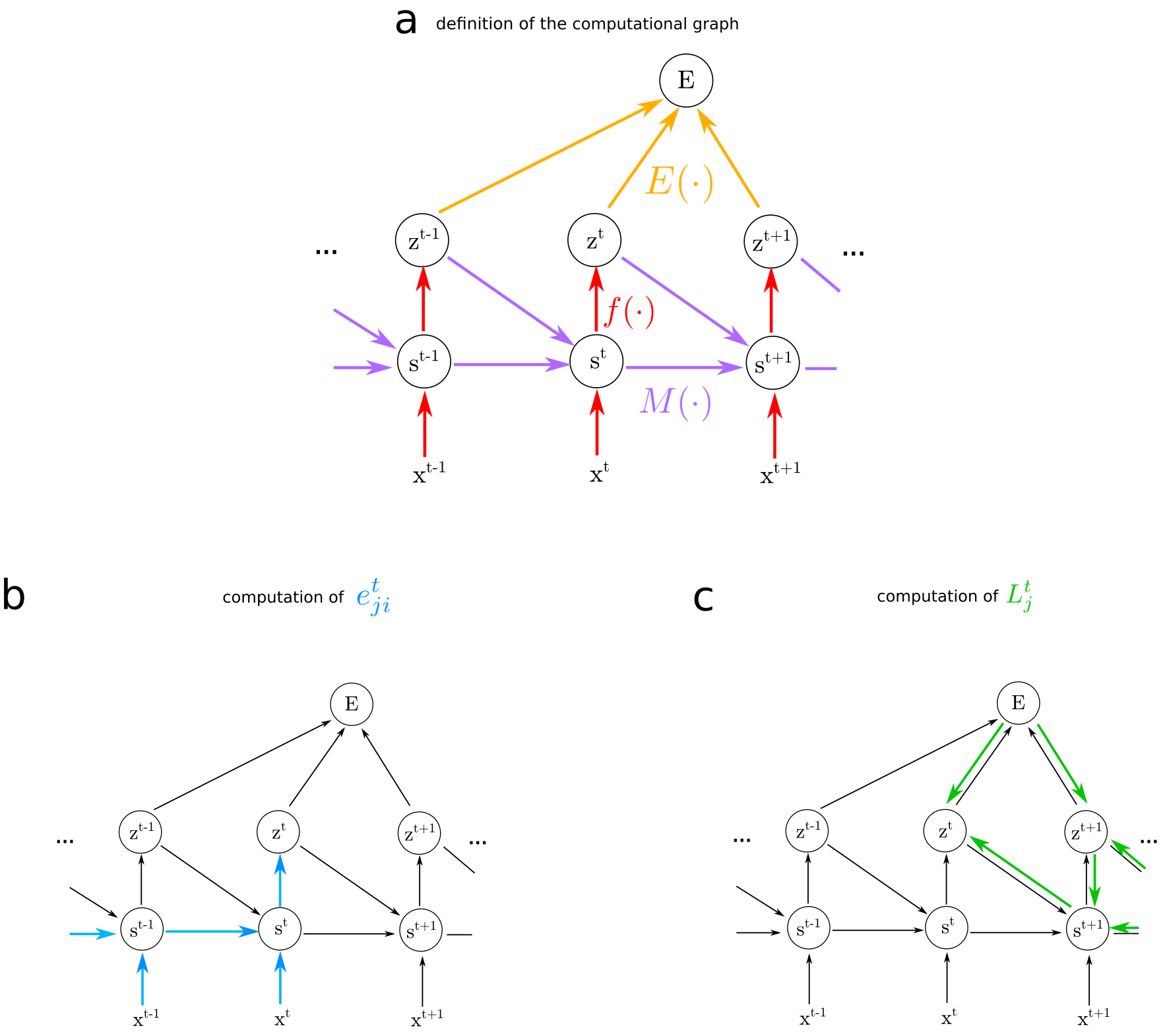}
    \caption{\textbf{Computational graph} \textbf{a)} Assumed mathematical dependencies between neuron states $\bt s_j^t$, neuron outputs $\bt z^t$, network inputs $\bt x^t$, and the network error $E$ through the mathematical functions $f(\cdot)$, $M(\cdot)$ and $E(\cdot)$ represented by coloured arrows.  \textbf{b)} The dependencies involved in the computation of the eligibility traces $e_{ji}^t$ are shown in blue.
    \textbf{c)} The dependencies involved in the computation of the learning signal $L_j^t$ are shown in green.
    }
    \label{fig:graph}
\end{figure}

\paragraph{Proof of factorization (equation~\eqref{eq:grad}):} We provide here the proof for equation~\eqref{eq:grad}, i.e., we show that the total derivative of the error function $E$ with respect to the parameters $\boldsymbol \theta$ can be written as a product of learning signals $L_j^t$ and eligibility traces $e_{ji}^t$. First, recall that in BPTT the error gradient is decomposed as (see equation (12) in \cite{werbos1990backpropagation}):

\begin{eqnarray}
	\frac{dE}{d \theta_{ji}} = \sum_t \frac{dE}{d \bt s_j^t} \cdot \frac{\partial \bt s_j^t}{\partial \theta_{ji}}~, \label{eq:bptt}
\end{eqnarray}
where $\frac{dE}{d \bt s_j^t}$ is the total derivative of the error $E$ with respect to the neuron states $\bt s_j^t$ at time step $t$. $\frac{dE}{d \bt s_j^t}$ can be expressed recursively as a function of the same derivative at the next time step $\frac{dE}{d \bt s_j^{t+1}}$ by applying the chain rule at the node $\bt s_j^t$ of the computational graph shown in Figure \ref{fig:graph}c:
\begin{eqnarray}
	\frac{d E}{d \bt s_j^t}
	& = & \frac{d E}{d z_j^t} \frac{\partial z_j^t}{\partial \bt s_j^t} + \frac{d E}{d \bt s_j^{t+1}} \frac{\partial \bt s_j^{t+1}}{\partial \bt s_j^t} \label{eq:backpropagation}\\
	& = &L_j^t \frac{\partial z_j^t}{\partial \textbf{s}_j^t} + \frac{d E}{d \bt s_j^{t+1}} D_j^t,  \label{eq:ex}
\end{eqnarray}
where we defined the {\em learning signal} for neuron $j$ at time $t$ as $L_j^t \overset{\text{def}}{=} \frac{dE}{d z_j^t} $.
The resulting recursive expansion ends at the last time step $T$, i.e., $\frac{dE}{d \bt s_j^{T+1}}=0$.
If one substitutes the recursive formula \eqref{eq:ex} into the definition of the error gradients \eqref{eq:bptt}, one gets:
\begin{eqnarray}
	\frac{dE}{d\theta_{ji}} & = & \sum_t \Bigg( L_j^t\frac{\partial z_j^t}{\partial \textbf{s}_j^t} + \frac{d E}{d \bt s_j^{t+1}} D_j^t \Bigg) \cdot \frac{\partial \bt s_j^t}{\partial \theta_{ji}} \\
	& = & \sum_{t} \bigg( L_j^t\frac{\partial z_j^t}{\partial \textbf{s}_j^t} + \big( L_j^{t + 1} \frac{\partial z_j^{t+1}}{\partial \textbf{s}_j^{t+1}} + (\cdots) D_j^{t + 1} \big) D_j^t \bigg) \cdot \frac{\partial \bt s_j^t}{\partial \theta_{ji}}~. \label{eq:dev}
\end{eqnarray}
The following equation is the main equation for understanding the transformation from BPTT into \textit{e-prop}. The key idea is to collect all terms which are multiplied with the learning signal $L_j^{t'}$ at a given time $t'$. These are only terms that concern events in the computation of neuron $j$ at time $t'$, and these do not depend on future errors or variables. Hence one can collect them conceptually into an internal eligibility trace $e_{ji}^t$ for neuron $j$ which can be computed autonomously within neuron $j$ in an online manner.

To this end, we write the term in parentheses in equation \eqref{eq:dev} into a second sum indexed by $t'$ and exchange the summation indices to pull out the learning signal $L_j^t$. This expresses the error gradient as a sum of learning signals $L_j^t$ multiplied by some factor indexed by $ji$, which implicitly defines what we call eligibility traces and eligibility vectors:
\begin{eqnarray}
	\frac{dE}{d\theta_{ji}}
	& = & \sum_{t} \sum_{t' \geq t} L_j^{t'} \frac{\partial z_j^{t'}}{\partial \textbf{s}_j^{t'}} D_j^{t'-1} \cdots D_j^t \cdot \frac{\partial \bt s_j^t}{\partial \theta_{ji}} \\
	& = & \sum_{t'} L_j^{t'} \frac{\partial z_j^{t'}}{\partial \textbf{s}_j^{t'}} \underbrace{\sum_{t \leq t'} D_j^{t'-1} \cdots D_j^t \cdot \frac{\partial \bt s_j^t}{\partial \theta_{ji}}}_{\stackrel{\text{def}}{=} \bt \bepsi_{ji}^{t'}}~. \label{eq:Dprod}
\end{eqnarray}
Here, we use the identity matrix for the $D^{t-1}_j \cdots D_j^t$ where $t'-1 < t$ . Finally, seeing that the eligibility vectors $\bt \bepsi_{ji}^{t'}$ can also be computed recursively as in equation \eqref{eq:elig-vector}, it proves the equation~\eqref{eq:grad}, given the definition of eligibility traces and learning signals in~\eqref{eq:elig-scalar} and~\eqref{eq:L}.

\paragraph{Leaky integrate-and-fire neuron model:} 
\label{sec:lif}

We define here the leaky integrate-and-fire (LIF) spiking neuron model, and exhibit the update rules that result from \textit{e-prop} for this model.
We consider LIF neurons simulated in discrete time. In this case the internal state $\bt s_j^t$ is one dimensional and contains only the membrane voltage $v_j^t$. The observable state $z_j^t \in \{0,1\}$ is binary, indicating a spike ($z_j^t=1$) or no spike ($z_j^t=0$) at time $t$. The dynamics of the LIF model is defined by the equations:
\begin{eqnarray}
v_j^{t+1} & = & \alpha v_j^t + \sum_{i \neq j} \theta_{ji}^\mathrm{rec} z_i^t + \sum_i \theta_{ji}^\mathrm{in} x_i^t - z_j^t v_\mathrm{th} \label{eq:lifv} \\
z_j^t &= & H\bigg( \frac{v_j^t - v_\mathrm{th}}{v_\mathrm{th}} \bigg) \label{eq:lifz},
\end{eqnarray}
where $x_i^t=1$ indicates a spike from the input neuron $i$ at time step $t$ ($x_i^t=0$ otherwise), $\theta_{ji}^\mathrm{rec}$ ($\theta_{ji}^\mathrm{in}$) is the synaptic weight from network (input) neuron $i$ to neuron $j$, and $H$ denotes the Heaviside step function. 
The decay factor $\alpha$ is given by $e^{-\delta t/\tau_m}$, where
$\delta t$ is the discrete time step (1 ms in our simulations) and
$\tau_m$ is the membrane time constant.  
Due to the term $-z_j^t v_\mathrm{th}$ in equation \eqref{eq:lifv}, the neurons membrane voltage is reset to a lower value after an output spike.

\textbf{Eligibility traces and error gradients:}
Considering the LIF model defined above, we derive the resulting eligibility traces and error gradients.
By definition of the model in equation \eqref{eq:lifv}, we have  $D_j^t = \frac{\partial v_j^{t+1}}{\partial v_j^{t}} = \alpha$ and $\frac{\partial v_j^{t}}{\partial \theta_{ji}} = z_i^{t-1}$. Therefore, using the definition of eligibility vectors in equation \eqref{eq:elig-vector}, one obtains a simple geometric series and one can write:
\begin{equation}
\bepsi_{ji}^{t+1} =  \sum_{t' \leq t} \alpha^{t - t'} z_i^{t'} \stackrel{\mathrm{def}}{=} \hat z_i^t~,
\label{eq:elig-vector-LIF}
\end{equation}
and the eligibility traces are written:
\begin{equation}
e_{ji}^{t+1} =  h_j^t \hat z_i^t~. \label{eq:elig-scalar-LIF}
\end{equation}

In other words, the eligibility vector is one dimensional and depends only on the presynaptic neuron $i$ and in fact, corresponds to the filtered presynaptic spike train.
To exhibit resulting the eligibility trace defined by equation \eqref{eq:elig-scalar}, this requires to compute the derivative $\frac{\partial z_j^t}{\partial v_j^t}$, which is ill-defined in the case of LIF neurons because it requires the derivative of the discontinuous function $H$.
As shown in \citep{bellec2018long}, we can alleviate this by using a pseudo-derivative $h_j^t$ in place of $\frac{\partial z_j^t}{\partial v_j^t}$, given by $h_j^t = \gamma \max \left( 0, 1 - | \frac{v_j^t - v_\mathrm{th}}{v_\mathrm{th}} |  \right)$ where $\gamma=0.3$ is a constant called dampening factor.
The gradient of the error with respect to a recurrent weight $\theta_{ji}^{\mathrm{rec}}$ thus takes on the following form, reminiscent of spike-timing dependent plasticity:
\begin{equation}
\frac{d E}{d \theta_{ji}^{\mathrm{rec}}} = \sum_t \frac{dE}{d z_j^t} h_j^t \hat{z}_i^{t-1}~. \label{eq:grad-LIF}
\end{equation}
A similar derivation leads to the gradient of the input weights. In fact, one just needs to substitute recurrent spikes $z_i^t$ by input spikes $x_i^t$. The gradient with respect to the output weights does not depend on the neuron model and is therefore detailed another paragraph (see equation \eqref{eq:e-prop1-methods-out}).

Until now refractory periods were not modeled to simplify the derivation. To introduce a simple model of refractory periods that is compliant with the theory, one can further assume that $z_j^t$ and $\frac{\partial z_j}{\partial \bt s_j}$ are fixed to $0$ for a short refractory period after each spike of neuron $j$. Outside of the refractory period the neuron dynamics are otherwise unchanged.

\paragraph{Leaky integrate-and-fire neurons with threshold adaptation:}
We derive the learning rule defined by \textit{e-prop} for a LIF neuron model with an adaptive threshold. For this model, the internal state is given by a two-dimensional vector $\bt s_j^t := (v_j^t, a_j^t)^T$, where $v_j^t$ denotes the membrane voltage as in the LIF model, and $a_j^t$ is a threshold adaptation variable. 
As for the LIF model above, the voltage dynamics is defined by equation \eqref{eq:lifv}. The spiking threshold $A_j^t$ at time $t$ is given by
\begin{equation}
A_j^t = v_\mathrm{th} + \beta a_j^t~ \label{eq:alifB},
\end{equation}
where $v_\mathrm{th}$ denotes the baseline-threshold. Output spikes are generated when the membrane voltage crosses the adaptive threshold $z_j^t = H\bigg( \frac{v_j^t - A_j^t}{v_\mathrm{th}} \bigg)$,
 and the threshold adaptation evolves according to
\begin{equation}
a_j^{t+1} = \rho a_j^t + H\bigg( \frac{v_j^t - A_j^t}{v_\mathrm{th}} \bigg) \label{eq:alifb}.
\end{equation}
The decay factor $\rho$ is given by $e^{-\delta t/\tau_a}$, where
$\delta t$ is the discrete time step (1 ms in our simulations) and
$\tau_a$ is the adaptation time constant.  In other words, the neuron's
threshold is increased with every output spike and decreases
exponentially back to the baseline threshold.

\textbf{Eligibility traces:}
Because of the extended state, we obtain one two-dimensional eligibility vector per synaptic weight: $\bt \bepsi_{ji}^t := (\epsilon_{ji,v}^t, \epsilon_{ji,a}^t)^T$ and the matrix $D_j^t$ is a $2\times 2$ matrix. On its diagonal one finds the terms $\frac{\partial v_j^{t+1}}{\partial v_j^t}=\alpha$ and $\frac{\partial a_j^{t+1}}{\partial a_j^t}=\rho - h_j^t\beta$.

Above and below the diagonal, one finds respectively $\frac{\partial v_j^{t+1}}{\partial a_j^{t}}=0$,$\frac{\partial a_j^{t+1}}{\partial v_j^{t}}= h_j^t$. One can finally compute the eligibility traces using its definition in equation \eqref{eq:elig-scalar}. The component of the eligibility vector associated with the voltage remains the same as in the LIF case and only depends on the presynaptic neuron: $\epsilon_{ji,v}^t = \hat z_i^{t-1}$. For the component associated with the adaptive threshold we find the following recursive update:
\begin{eqnarray}
	\epsilon_{ji,a}^{t+1} & = & h_j^{t} \hat z_i^{t-1} + (\rho - h_j^t \beta) \epsilon_{ji,a}^t~, \label{eq:elig-vector-ALIF} 
\end{eqnarray}
and this results in an eligibility trace of the form:

\begin{eqnarray}
	e_{ji}^{t+1} & = & h_j^t \bigg( \hat z_i^{t-1} - \beta \epsilon_{ji,a}^t \bigg). \label{eq:elig-scalar-ALIF}
\end{eqnarray}
This results in the following equation for the gradient of recurrent weights:
\begin{eqnarray}
	\frac{dE}{d\theta_{ji}^\mathrm{rec}} = \sum_t \frac{dE}{dz_j^t} h_j^t \bigg( \hat z_i^{t-1} - \beta \epsilon_{ji,a}^t \bigg)~. \label{eq:grad-ALIF}
\end{eqnarray}
The eligibility trace for input weights is again obtained by replacing recurrent spikes $z_i^t$ with input spikes $x_i^t$.

\paragraph{Artificial neuron models:}
The dynamics of recurrent artificial neural networks is usually given by
$s_j^t=\alpha s_j^{t-1}+\sum_i\theta_{ji}^{\mathrm{rec}}z_i^{t-1}+\sum_i\theta_{ji}^{\mathrm{in}} x_j^t$ with $z_j^t=\sigma(s_j^t)$, where $\sigma: \RR \rightarrow \RR$ is some activation function (often a sigmoidal function in RNNs).
We call the first term the leak term in analogy with LIF models. For $\alpha=0$ this term disappears, leading to the arguably most basic RNN model. If $\alpha=1$, it models a recurrent version of residual networks.

\textbf{Eligibility traces:}
For such model, one finds that $D_j^t=\alpha$ and the eligibility traces are equal to $e_{ji}^t= h_j^t \hat{z}_i^{t-1}$ with $\hat{z}_i^t \eqdef \sum_{t'\leq t} z_i^{t'} \alpha^{t-t'}$. The resulting \textit{e-prop} update is written as follows (with $ \sigma'$ the derivative of the activation function):
\begin{equation}
\frac{dE}{d\theta_{ji}^\mathrm{rec}} = \sum_t \frac{dE}{dz_j^t} \sigma'(s_j^t) \hat z_i^{t-1}. \label{eq:ann-elig}
\end{equation}
Although simple, this derivation provides insight in the relation between BPTT and \textit{e-prop}.
If the neuron model does not have neuron specific dynamics $\alpha=0$, the factorization of \textit{e-prop} is obsolete in the sense that the eligibility trace does not propagate any information from a time step to the next $D_j^t=0$. Thus, one sees that \textit{e-prop} is most beneficial for models with rich internal neural dynamics.

\paragraph{LSTM:}
For LSTM units, \citep{hochreiter_long_1997} the internal state of the unit is the content of the memory cell and is denoted by $c_j^t$, the observable state is denoted by $h_j^t$. 
One defines the network dynamics that involves the usual input, forget and output gates (denoted by  $\mathscr{i}_j^t$, $\mathscr{f}_j^t$, and $\mathscr{o}_j^t$) and the cell state candidate $\widetilde{c}_j^t$ as follows (we ignore biases for simplicity):
\begin{eqnarray}
 \mathscr{i}_j^t & = & \sigma \big( \sum_i \theta_{ji}^{\mathrm{rec},\mathscr{i}} h_i^{t-1} + \sum_i \theta_{ji}^{\mathrm{in},\mathscr{i}} x_i^t \big) \\
 \mathscr{f}_j^t & = & \sigma \big( \sum_i \theta_{ji}^{\mathrm{rec},\mathscr{f}} h_i^{t-1} + \sum_i \theta_{ji}^{\mathrm{in},\mathscr{f}} x_i^t \big) \\
 \mathscr{o}_j^t & = & \sigma \big( \sum_i \theta_{ji}^{\mathrm{rec},\mathscr{o}} h_i^{t-1} + \sum_i \theta_{ji}^{\mathrm{in},\mathscr{o}} x_i^t \big) \\
 \widetilde{c}_j^t & = & \operatorname{tanh} \big( \sum_i \theta_{ji}^{\mathrm{rec},c} h_i^{t-1} + \sum_i \theta_{ji}^{\mathrm{in},c} x_i^t \big).
\end{eqnarray}
Using those intermediate variables as notation short-cuts, one can now write the update of the states of LSTM units in a form that we can relate to \textit{e-prop}:
\begin{eqnarray}
 c_j^{t} & = & M(c_j^{t-1},\bt h^{t-1}, \boldsymbol \theta) = \mathscr{f}_j^{t} c_j^{t-1} + \mathscr{i}_j^{t} \widetilde{c}_j^{t} \\
 h_j^{t} & = & f(c_j^{t}, \bt h^{t-1}, \boldsymbol \theta) = \mathscr{o}_j^{t} c_j^{t}.
\end{eqnarray}

\textbf{Eligibility traces:}
There is one difference between LSTMs and the previous neuron models used for \textit{e-prop}: the function $f$ depends now on the previous observable state $\bt h^t$ and the parameters through the output gate $\mathscr{o}_j^t$.
However the derivation of the gradients $\frac{d E}{d \theta_{ji}}$ in the paragraph ``Proof of factorization'' is still valid for deriving the gradients with respect to the parameters of the input gate, forget gate and cell state candidate. 
For these parameters we apply the general theory as follows.
We compute $D_j^t=\frac{\partial c_j^{t+1}}{\partial c_j^{t}} = \mathscr{f}_j^{t}$ and for each variable $\theta_{ji}^{A,B}$ with $A$ being either ``$\mathrm{in}$'' or ``$\mathrm{rec}$'' and $B$ being $\mathscr{i},\mathscr{f}$, or $c$, we compute a set of eligibility traces. If we take the example the recurrent weights for the input gate $\theta_{ji}^{\mathrm{rec},\mathscr{i}}$, the eligibility vectors are updated according to:
\begin{eqnarray}
\bepsi_{ji}^{\mathrm{rec},\mathscr{i},t} = \mathscr{f}_j^{t-1} \bepsi_{ji}^{\mathrm{rec},\mathscr{i},t-1} + \widetilde{c}_j^t {\mathscr{i}}_j^t (1 - {\mathscr{i}}_j^t) h_i^t~, \label{eq:elig-vector-lstm}
\end{eqnarray}
the eligibility traces are then written:
\begin{eqnarray}
e_{ji}^{\mathrm{rec},\mathscr{i},t} = \mathscr{o}_j^t \bepsi_{ji}^{\mathrm{rec},\mathscr{i},t} , \label{eq:elig-scalar-lstm}
\end{eqnarray}
and the gradients are of the form
\begin{eqnarray}
\frac{dE}{d \theta_{ji}^{\mathrm{rec},\mathscr{i}}} = \sum_t \frac{dE}{dh_j^t} \mathscr{o}_j^t \bepsi_{ji}^{\mathrm{rec},\mathscr{i},t}~. \label{eq:grad-lstm}
\end{eqnarray}

For the parameters $\theta_{ji}^{\mathrm{rec},\mathscr{o}}$ of the output gate which take part in the function $f$, we need to derive the gradients in a different manner.
As we still assume that $E(\bt h^1, \dots, \bt h^T)$ depends on the observable state only, we can follow the derivation of BPTT with a formula analogous to \eqref{eq:bptt}. This results in a gradient expression involving local terms and the same learning signal as used for other parameters.
Writing $\frac{\partial f}{\partial \theta_{ji}^{\mathrm{rec},\mathscr{o}}}$ as $\frac{\partial h_j^{t}}{\partial \theta_{ji}^{\mathrm{rec},\mathscr{o}}}$ the gradient update for the output gate takes the form:
\begin{equation}
\frac{dE}{d \theta_{ji}^{\mathrm{rec},\mathscr{o}}}
= \sum_t \frac{dE}{d h_j^t} \frac{\partial h_j^{t}}{\partial \theta_{ji}^{\mathrm{rec},\mathscr{o}}}
= \sum_t \frac{dE}{d h_j^t} c_j^t \mathscr{o}_j^{t} (1 - \mathscr{o}_j^{t}) h_i^{t-1}.
\end{equation}

\subsection*{\textit{E-prop 1}}
\label{sec:methods_e-prop1}
\textit{E-prop 1} follows the general \textit{e-prop} framework and applies to all the models above.
Its specificity is the choice of learning signal.
In this first variant, we make two approximations: future errors are ignored so that one can compute the learning signal in real-time, and learning signals are fed back with random connections. The specific realizations of the two approximations are discussed independently in the following and implementation details are provided.

\textbf{Ignoring future errors:}
The first approximation is to focus on the error at the present time $t$ and ignore dependencies on future errors in the computation of the total derivative $\frac{d E}{d z_j^t}$.
Using the chain rule, this total derivative expands as $\frac{d E}{d z_j^t}=\frac{\partial E}{\partial z_j^t} + \frac{d E}{d \bt s_j^{t+1}} \frac{\partial \bt s_j^{t+1}}{\partial z_j^t}$, and neglecting future errors means that we ignore the second term of this sum. As a result the total derivative $\frac{d E}{d z_j^t}$ is replaced by the partial derivative $\frac{\partial E}{\partial z_j^t}$ in equation \eqref{eq:L}.

\textbf{Synaptic weight updates under \textit{e-prop 1}:}
Usually, the output of an RNN is given by the output of a set of readout neurons which receive input from network neurons, weighted by synaptic weights $\theta_{kj}^{\mathrm{out}}$. In the case of an RSNN, in order to be able to generate non-spiking outputs, readouts are modeled as leaky artificial neurons. More precisely, the output of readout $k$ at time $t$ is given by
\begin{equation}
  y_k^{t}=\kappa y_k^{t-1} + \sum_j \theta_{kj}^{\mathrm{out}} z_j^{t} + b_k^{\mathrm{out}},
\end{equation}
where $\kappa \in [0,1]$ defines the leak and $b_k^{\mathrm{out}}$ denotes the readout bias. The leak factor $\kappa$ is given by $e^{-\delta t/\tau_{out}}$, where
$\delta t$ is the discrete time step and $\tau_{out}$ is the membrane time constant). In the following derivation of weight updates under \textit{e-prop 1}, we assume such readout neurons. Additionally, we assume that the error function is given by the mean squared error $E=\frac{1}{2}\sum_{t,k}(y_k^t-y_k^{*,t})^2$ with $y_k^{*,t}$ being the target output at time $t$ (see the following paragraph on ``classification'' when the cross entropy error is considered).

In this case, the partial derivative $\frac{\partial E}{\partial z_i^t}$ has the form:
\begin{equation}
\frac{\partial E}{\partial z_j^t}= \theta_{kj}^{\mathrm{out}}\sum_{t' \geq t}(y_k^{t'} - y_k^{*,t'}) \kappa^{t' - t}.  \label{eq:dEdz-mse}
\end{equation}
This seemingly poses a problem for a biologically plausible learning rule, because the partial derivative is a weighted sum over the future. This issue can however easily be solved as we show below. Using equation \eqref{eq:grad} for the particular case of \textit{e-prop 1}, we insert $\frac{\partial E}{\partial z_j^t}$ in-place of the total derivative $\frac{d E}{d z_j^t}$ which leads to an estimation $\widehat{\frac{d E}{d \theta_{ji}}}$ of the true gradient given by:
\begin{eqnarray}
\widehat{\frac{d E}{d \theta_{ji}}} & = & \sum_t \frac{\partial E}{\partial z_j^t} e_{ji}^t \label{eq:e-prop1-methods-general} \\
 & = & \sum_{k,t} \theta_{kj}^{\mathrm{out}} \sum_{t' \geq t} (y_k^t-y_k^{*,t})\kappa^{t' - t} e_{ji}^t \\
 & = & \sum_{k,t'} \theta_{kj}^{\mathrm{out}} (y_k^{t'}-y_k^{*,{t'}})  \sum_{t \leq t'}\kappa^{t' - t} e_{ji}^t~, \label{eq:e-prop1-methods}
\end{eqnarray}
where we inverted sum indices in the last line.
The second sum indexed by $t$ is now over previous events that can be computed in real time, it computes a filtered copy of the eligibility trace $e_{ji}^t$. With this additional filtering of the eligibility trace with a time constant equal to that of the temporal averaging of the readout neuron, we see that \textit{e-prop 1} takes into account the latency between an event at time $t'$ and its impact of later errors at time $t$ within the averaging time window of the readout. However, note that this time constant is a few tens of milliseconds in our experiments which is negligible in comparison to the decay of the eligibility traces of adaptive neurons which are one to two orders of magnitude larger (see Figure \ref{fig:store_recall}). 

Equation \eqref{eq:e-prop1-methods} holds for any neuron model. In the case of LIF neurons, the eligibility traces are given by equation \eqref{eq:elig-scalar-LIF}, and one obtains the final expression of the error gradients after substituting these expressions in \eqref{eq:e-prop1-methods}. Implementing weight updates with gradient descent and learning rate $\eta$, the updates of the recurrent weights are given by
\begin{eqnarray}
\Delta \theta_{ji}^{\mathrm{rec}} & = \eta & \sum_t \big( \sum_{k} \theta_{kj}^{\mathrm{out}} (y_k^{*,{t}}-y_k^{t}) \big)  \sum_{t' \leq t} \kappa^{t - t'} h_j^{t'} \hat{z}_{i}^{t'-1}~. \label{eq:e-prop1-methods-sym-feedback-LIF}
\end{eqnarray}
When the neurons are equipped with adaptive thresholds as in LSNNs, one replaces the eligibility traces with their corresponding definitions.
It results that an additional term $\epsilon_{ji,a}^t$ as defined in equation \eqref{eq:elig-vector-ALIF} is introduced in the weight update:
\begin{eqnarray}
\Delta \theta_{ji}^{\mathrm{rec}} & = \eta & \sum_t \big( \sum_{k} \theta_{kj}^{\mathrm{out}} (y_k^{*,{t}}-y_k^{t}) \big)  \sum_{t' \leq t}\kappa^{t - t'} h_j^{t'} \left( \hat{z}_{i}^{t'-1} - \beta \epsilon_{ji,a}^{t'}\right)~. \label{eq:e-prop1-methods-sym-feedback-ALIF}
\end{eqnarray}
Both equations \eqref{eq:e-prop1-methods-sym-feedback-LIF} and \eqref{eq:e-prop1-methods-sym-feedback-ALIF} can be derived similarly for the input weights $\theta_{ji}^{\mathrm{in}}$, and it results in the same learning rule with the only difference that $\hat{z}_{i}^{t'-1}$ is replaced by a trace of the spikes $x_i^t$ of input neuron $i$. For the output connections the gradient $\frac{d E}{d \theta_{kj}^{\mathrm{out}}}$ can be derived as for isolated linear readout neurons and it does not need to rely on the theory of \textit{e-prop}. The resulting weight update is:
\begin{eqnarray}
\Delta \theta_{kj}^{\mathrm{out}} & = \eta & \sum_t (y_k^{*,{t}}-y_k^{t}) \sum_{t' \leq t}\kappa^{t - t'} z_j^{t'}~. \label{eq:e-prop1-methods-out}
\end{eqnarray}

\textbf{Random feed-back matrices:}
According to equations \eqref{eq:e-prop1-methods-sym-feedback-LIF} and \eqref{eq:e-prop1-methods-sym-feedback-ALIF}, the signed error signal from readout $k$ communicated to neuron $j$ has to be weighted with $\theta_{kj}^{\mathrm{out}}$. That is, the synaptic efficacies of the feedback synapses have to equal those of the feed-forward synapses.  This general property of backpropagation-based algorithms is a problematic assumption for biological circuits. It has been shown however in \citep{samadi2017deep,nokland2016direct} that for many tasks, an approximation where the feedback weights are chosen randomly works well. We adopt this approximation in \textit{e-prop 1}.
Therefore we replace in equations \eqref{eq:e-prop1-methods-sym-feedback-LIF} and \eqref{eq:e-prop1-methods-sym-feedback-ALIF} the weights $\theta_{kj}^{\mathrm{out}}$ by fixed random values $B_{jk}^{\mathrm{random}}$. For a LIF neuron the learning rule \eqref{eq:e-prop1-methods-sym-feedback-LIF} becomes with random feedback weights:
\begin{eqnarray}
\Delta \theta_{ji}^{\mathrm{rec}} & = \eta & \sum_t \big( \sum_{k} B_{jk}^{\mathrm{random}} (y_k^{*,{t}}-y_k^{t}) \big)  \sum_{t' \leq t}\kappa^{t - t'} h_j^{t'} \hat{z}_{i}^{t'-1}~. \label{eq:e-prop1-methods-LIF}
\end{eqnarray}
For an adaptive LIF neuron the learning rule \eqref{eq:e-prop1-methods-sym-feedback-ALIF} becomes with random feed-back weights:
\begin{eqnarray}
\Delta \theta_{ji}^{\mathrm{rec}} & = \eta & \sum_t \big( \sum_{k} B_{jk}^{\mathrm{random}} (y_k^{*,{t}}-y_k^{t}) \big)  \sum_{t' \leq t}\kappa^{t - t'} h_j^{t'} \left( \hat{z}_{i}^{t'-1} - \beta \epsilon_{ji,a}^{t'}\right) ~. \label{eq:e-prop1-methods-ALIF}
\end{eqnarray}

\textbf{Synaptic weight updates under \textit{e-prop 1} for classification:}
For the classification tasks solved with \textit{e-prop 1}, we consider one readout neuron $y_k^t$ per output class, and the network output at time $t$ corresponds to the readout with highest voltage.
To train the recurrent networks in this setup, we replace the mean squared error by the the cross entropy error ${E=-\sum_{t,k}\pi_k^{*,t}\log \pi_k^{t}}$ where the target categories are provided in the form of a one-hot-encoded vector $\pi_k^{*,t}$. On the other hand, the output class distribution predicted by the network is given as ${\pi_k^{t}=\operatorname{softmax}(y_k^t)= \exp(y_k^t) / \sum_{k'} \exp(y_{k'}^t)}$. To derive the modified learning rule that results from this error function $E$, we replace $\frac{\partial E}{\partial z_i}$ of equation \eqref{eq:dEdz-mse} with the corresponding gradient:
\begin{equation}
\frac{\partial E}{\partial z_j^t}= \theta_{kj}^{\mathrm{out}}\sum_{t' \geq t}(\pi_k^{t'} - \pi_k^{*,{t'}}) \kappa^{t' - t}.  \label{eq:dEdz-cee}
\end{equation}
Following otherwise the same derivation as previously it results that the weight update of \textit{e-prop 1} previously written in equation \eqref{eq:e-prop1-methods-LIF} becomes for a LIF neuron:
\begin{eqnarray}
\Delta \theta_{ji}^{\mathrm{rec}} & = \eta & \sum_t \big( \sum_{k} B_{jk}^{\mathrm{random}} (\pi_k^{*,{t}}-\pi_k^{t}) \big)  \sum_{t' \leq t}\kappa^{t - t'} h_j^{t'} \hat{z}_{i}^{t'-1}~. \label{eq:e-prop1-methods-ce-LIF}
\end{eqnarray}
Similarly, for the weight update of the output connections, the only difference between the update rules for regression and classification is that the output $y_k^t$ and the target $y_k^{*,t}$ are respectively replaced by $\pi_k^{t}$ and $\pi_k^{*,{t}}$:
\begin{eqnarray}
\Delta \theta_{kj}^{\mathrm{out}} & = \eta & \sum_t (\pi_k^{*,{t}}-\pi_k^{t}) \sum_{t' \leq t}\kappa^{t - t'} z_j^{t'}~. \label{eq:e-prop1-methods-ce-out}
\end{eqnarray}

\textbf{Firing rate regularization:}
To ensure that the network computes with low firing rates, we add a regularization term $E_{\mathrm{reg}}$ to the error function $E$.
This regularization term has the form: 
\begin{equation}
   E_{\mathrm{reg}} = \sum_j \left( f_j^{\mathrm{av}} - f^{\mathrm{target}} \right)^2~, \label{eq:regularization}
\end{equation}
where $f^{\mathrm{target}}$ is a target firing rate and $f_j^{\mathrm{av}}=\frac{\delta t}{n_{\mathrm{trials}} T}\sum_{t,k} z_j^t$ is the firing rate of neuron $j$ averaged over the $T$ time steps and the $n_{\mathrm{trials}}$ trials separating each weight update.
To compute the weight update that implements this regularization, we follow a similar derivation as detailed previously for the mean square error.
Instead of equation \eqref{eq:dEdz-mse}, the partial derivative has now the form:
\begin{equation}
\frac{\partial E_{\mathrm{reg}}}{\partial z_j^t}= \frac{\delta t}{n_{\mathrm{trials}} T} \left(f_j^{\mathrm{av}}-f^{\mathrm{target}} \right). \label{eq:dEdz-reg}
\end{equation}
Inserting this expression into the equation \eqref{eq:e-prop1-methods-general}, and choosing the special case of a LIF neurons,  it results that the weight update that implements the regularization is written:
\begin{eqnarray}
\Delta \theta_{ji}^{\mathrm{rec}} & = \eta &  \sum_t \frac{\delta t}{n_{\mathrm{trials}} T} \left(f^{\mathrm{target}}- f_j^{\mathrm{av}} \right) h_j^{t} \hat{z}_{i}^{t-1}~. \label{eq:e-prop1-methods-LIF-reg}
\end{eqnarray}
The same learning rule is also applied to the input weights $\Delta \theta_{ji}^{\mathrm{in}}$.
This weight update is performed simultaneously with the weight update exhibited in equation \eqref{eq:e-prop1-methods-LIF} which optimizes the main error function $E$. For other neuron models such as adaptive LIF neurons, the equation \eqref{eq:e-prop1-methods-LIF-reg} has to be updated accordingly to the appropriate definition of the eligibility traces.

\subsection*{Details to simulations for \textit{E-prop 1}}
\label{sec:methods_e-prop1_sim}

\paragraph{General simulation and neuron parameters:}
In all simulations of this article, networks were simulated in discrete time with a simulation time step of 1 ms. Synapses had a uniform transmission delay of $1$ ms. Synaptic weights of spiking neural networks were initialized as in \citep{bellec2018long}. 

\paragraph{Implementation of the optimization algorithm:}
A dampening factor of $\gamma = 0.3$ for the pseudo-derivative of the
spiking function was used in all simulations of this article. The
weights were kept constant for $n_{batch}$ independent trials
(specified for individual simulations below), and the gradients were
cumulated additively. After collecting the gradients, the weights were
updated using the Adam algorithm \citep{kingma2014adam}.  For all
simulations of \textit{e-prop 1}, the gradients were computed according to
equation \eqref{eq:e-prop1-methods}.


\paragraph{Integration of the ``Clopath rule'' in \textit{e-prop 1}:}
We replaced the presynpatic and postsynaptic factors of equation \eqref{eq:elig-scalar-LIF} with the model of long term potentiation defined in \cite{clopath2010connectivity} and fitted to data in the same paper. The learning rule referred as ``Clopath rule'' in our experiments differ from \textit{e-prop 1} by the replacement of the pseudo derivative $h_j^t$ by another non linear function of the post synpatic voltage. In comparison to equation \eqref{eq:grad-LIF} the computation of the error gradient becomes:
\begin{eqnarray}
\frac{dE}{d\theta_{ji}^{\mathrm{rec}}} & = & \sum_t \frac{dE}{dz_j^t} [v_j^{t} - v_\mathrm{th}^+]^+ [\hat{v}_j^{t} - v_\mathrm{th}^-]^+ \hat{z}_{i}^{t-1}~, \label{eq:e-prop1-clopath}
\end{eqnarray}
where $\hat{v}_j^{t}$ is an exponential trace of the post synaptic membrane potential with time constant $10$~ms and $[\cdot]^+$ is the rectified linear function. The time constant of $\hat{v}_j^{t}$ was chosen to match their data. The thresholds $v_\mathrm{th}^-$ and $v_\mathrm{th}^+$ were $\frac{v_\mathrm{th}}{4}$ and $0$ respectively. All other implementation details remained otherwise unchanged between the default implementation of \textit{eprop 1} and this variant of the algorithm.

\paragraph{Pattern generation task 1.1:}
The three target sequences had a duration of $1000$ ms and were given by the sum of four sinusoids for each sequence with fixed frequencies of 1 Hz, 2 Hz, 3 Hz, and 5 Hz. The amplitude of each sinusoidal component was drawn from a uniform distribution over the interval $[0.5, 2]$. Each component was also randomly phase-shifted with a phase sampled uniformly in the interval $[0, 2\pi)$.

The network consisted of 600 all-to-all recurrently connected LIF neurons (no adaptive thresholds). The neurons had a membrane time constant of $\tau_m = 20$ ms and a refractory period of $5$ ms. The firing threshold was set to $v_\mathrm{th} = 0.61$. The network outputs were provided by the membrane potential of three readout neurons with a time constant $\tau_{out} = 20$ ms.
The network received input from 20 input neurons, divided into 5 groups, which indicated the current phase of the target sequence similar to \citep{nicola2017supervised}. Neurons in group $i \in \{0, 4\}$ produced 100 Hz regular spike trains during the time interval $[200\cdot i, 200 \cdot i+200)$ ms and were silent at other times.

A single learning trial consisted of a simulation of the network for $1000$ ms, i.e., the time to produce the target pattern at the output. The input, recurrent, and output weights of the network were trained for $1000$ iterations with the Adam algorithm, a learning rate of $0.003$ and the default hyperparameters \citep{kingma2014adam}. After every $100$ iterations, the learning rate was decayed with a multiplicative factor of $0.7$. A batch size of a single trial was used for training. To avoid an implausibly high firing rate, a regularization term was added to the loss function, that keeps the neurons closer to a target firing rate of $10$ Hz. The regularization loss was given by the mean squared error (mse) between the mean firing rate of all neurons over a batch and the target rate. This loss was multiplied with the factor $0.5$ and added with the target-mse to obtain the total loss to be optimized.

The comparison algorithms in Figure \ref{fig:partials}d,e were implemented as follows. When training with \textit{e-prop 1}, the random feedback weights $B^{\mathrm{random}}$ were generated from a Gaussian distribution with mean $0$ and variance $\frac{1}{n}$, where $n$ is the number of network neurons. For the performance of the global error signal, \textit{e-prop 1} was used, but the random feedback matrix was replaced by a matrix where all entries had the value $\frac{1}{\sqrt{n}}$. As a second baseline a network without recurrent connections was trained with \textit{e-prop 1} (``No rec. conn.'' in panel d). We further considered variants of \textit{e-prop 1} where we sampled independent feedback matrices for every $1$ or $20$ ms window (``$1$ ms'' and ``$20$ ms'' in panels d and e). Note that the same sequence of feedback matrices had to be used in every learning trials. We also compared to BPTT, where the Adam algorithm was used with the same meta-parameters as used for \textit{e-prop 1}.

\paragraph{Store-recall task 1.2:}
The store-recall task is described in Results. Each learning trial consisted of a $2400$ ms network simulation. We used a recurrent LSNN network consisting of $10$ standard LIF neurons and $10$ LIF neurons with adaptive thresholds. All neurons had a membrane time constant of $\tau_{m} = 20$ ms and a baseline threshold of $v_\mathrm{th} = 0.5$. Adaptive neurons had a threshold increase constant of $\beta = 0.03$ and a threshold adaptation time constant of $\tau_{a} = 1200$ ms. A refractory period of $5$~ms was used. The input, recurrent and output weights of the network were trained with a learning rate of $0.01$ and the Adam algorithm the default hyperparameters \citep{kingma2014adam}. Training was stopped when a misclassification rate below $0.05$ was reached.  After $100$ iterations, the learning rate was decayed with a multiplicative factor of $0.3$. The distribution of the random feedback weights $B^{\mathrm{random}}$ was generated from normal distribution with mean $0$ and variance $\frac{1}{n}$, where $n$ is the number of recurrent neurons. A batch size of $128$ trials was used.

In Figure \ref{fig:store_recall}b, we quantified the information content of eligibility traces at training iteration 25, 75, and 200 in this task. After the predefined number of training iterations, we performed test simulations where we provided only a store command to the network and simulated the network up to $6000$ ms after this store. A linear classifier (one for a time window of $100$ ms, at every multiple of $50$ ms) was then trained to predict the stored bit from the value of the eligibility traces at that time. For this purpose we used logistic regression with a squared regularizer on the weights. We used $150$ different simulations to train the classifiers and evaluated the decoding accuracy, as shown in Figure~\ref{fig:store_recall}b, on $50$ separate simulations.

\paragraph{Speech recognition task 1.3:}
We followed the same task setup as in \citep{greff2017lstm, graves2005framewise}.
The TIMIT dataset was split according to Halberstadt \citep{halberstadt} into a training, validation, and test set with 3696, 400, and 192 sequences respectively.
The networks received preprocessed audio at the input.
Preprocessing of audio input consisted of the following steps:
computation of $13$ Mel Frequency Cepstral Coefficients (MFCCs) with frame size $10$ ms on input window of $25$ ms,
computation of the first and the second derivatives of MFCCs,
concatenation of all computed factors to $39$ input channels.
Input channels were mapped to the range $[0, 1]$ according to the minumum/maximum values in the training set.
These continuous values were used directly as inputs $x_i^t$ in equation \eqref{eq:lifv}.

To tackle this rather demanding benchmark task, we used a bi-directional network architecture \citep{graves2005framewise}, that is, the standard LSNN network was appended by a second network which recieved the input sequence in reverse order. 
A bi-directional LSNN (300 LIF neurons and 100 adaptive LIF neurons per direction) was trained with different training algorithms.
Unlike in task 1.1, the random feedback weights $B^{\mathrm{random}}$ were generated with a variance of $1$ instead of $\frac{1}{n}$ as we observed that it resulted in better performances for this task.

With LSNNs we first ran a simple $8$ point grid search over the firing threshold hyperparameter $v_\mathrm{th}$.
The best performing value for threshold was then used to produce the LSNN results (see Figure~\ref{fig:store_recall}c).
For the strong baseline we include the result of LSTMs applied to the same task \citep{greff2017lstm},
where the hyperparameters were optimized using random search for 200 trials over the following hyperparameters:
number of LSTM blocks per hidden layer, learning rate, momentum, momentum type, gradient clipping, and standard deviation of Gaussian input noise.
In \citep{greff2017lstm} the mean test accuracy of 10\% best performing hyperparameter settings (out of 200) is $0.704$.

Every input step which represents the $10$ ms preprocessed audio frame is fed to the LSNN network for $5$ consecutive $1$ ms steps.
All neurons had a membrane time constant of $\tau_m = 20$ ms and a refractory period of $2$ ms.
Adaptive neurons had $\beta = 1.8$ and an adaptation time constant of $\tau_a = 200$ ms.
We used $61$ readout neurons, one for each class of the TIMIT dataset. A softmax was applied to their output, which was used to compute the cross entropy error against the target label.
Networks were trained using Adam with the default hyperparameters \citep{kingma2014adam} except $\epsilon_\mathrm{Adam} = 10^{-5}$.
The learning rate was fixed to $0.01$ during training.
We used a batch size of $32$ and the membrane time constant of the output neurons was $3$ ms.
Regularizaion of the network firing activity was applied as in Task 1.1.



\subsection*{\textit{E-prop 2}} 
\label{sec:methods_e-prop2}
In \textit{e-prop 2}, the learning signals are computed in a separate error module.
In order to distinguish the error module from the main network, we define a separate internal state vector for each neuron $j$ in the error module $\bm \sigma_j^t$ and network dynamics $\bm \sigma_j^t = M_e(\bm \sigma_j^{t-1}, \bm \zeta^{t-1}, \bm \xi^t, \bm \Psi)$ for it. Here, $\bm \zeta^{t-1}$ is the vector of neuron outputs in the error module at time $t$, and synaptic weights are denoted by $\bm \Psi$. The inputs to the error module are written as: $\bm \xi^t = (\bm x^t, \bm z^t, \bm y^{*,t})$ with $\bm y^{*,t}$ denoting the target signal for the network at time $t$. Note that the target signal is not necessarily the target output of the network, but can be more generally a target state vector of some controlled system. For example, the target signal in task 2.1 is the target position of the tip of an arm at time $t$, while the outputs of the network define the angular velocities of arm joints. 

The error module produces at each time $t$ a learning signal $\hat L_j^t$ for each neuron $j$ of the network, which were computed according to: 
\begin{equation}
 \hat L_j^t = \alpha_e \hat L_j^{t-1} + \sum_i \Psi{ji}^\mathrm{out} \zeta_i^t~,
\end{equation}
where the constant $\alpha_e$ defines the decay of the resulting learning signal, e.g. the concentration of a neuromodulator.

\textbf{Synaptic weight updates under e-prop 2: }
The task of the error module is to compute approximations to the true learning in equation~\eqref{eq:L}. Therefore, in comparison to equation~\eqref{eq:elig-scalar-LIF}, we obtain an estimation of the true error gradients $\frac{dE}{d \theta_{ji}^\mathrm{rec}}$ given as ${\widehat{\frac{dE}{d \theta_{ji}^\mathrm{rec}}} = \sum_t \hat L_j^t h_j^t \hat{z}_i^{t-1}}$, which in turn leads to an update rule for the synaptic weights using a fixed learning rate $\eta$:
\begin{equation}
	\Delta \theta^\mathrm{rec}_{ji} = - \eta \sum_t \hat L_j^t h_j^t \hat z_i^{t-1}
\end{equation}
Similarly, the update rule for input weights is obtained by replacing $\hat z_i^{t-1}$ in favor of $\hat x_i^{t-1}$.

In the experiments regarding \textit{e-prop 2}, input and recurrent weights were updated a single time in the inner loop of L2L according to $\bm \theta_\mathrm{test} = \bm \theta_\mathrm{init} + \Delta \bm \theta$, whereas output weights were kept constant.

\paragraph{Target movement task 2.1:}
In this task, the two network outputs are interpreted as angular velocities $\dot \phi_1$ and $ \dot \phi_2$ and are applied to the joints of a simple arm model.
The configuration of the arm model at time $t$ is described by the angles $\phi_1^t$ and $\phi_2^t$ of the two joints measured against the horizontal and the first leg of the arm respectively, see Figure~\ref{fig:one-shot}c. For given angles, the position $\bm y^t = (x^t, y^t)$ of the tip of the arm in Euclidean space is given by ${x^t = l \cos(\phi_1^t) + l \cos(\phi_1^t + \phi_2^t)}$ and $y^t = l \sin(\phi_1^t) + l \sin(\phi_1^t + \phi_2^t)$. Angles were computed by discrete integration over time: $\phi_i^t = \sum_{t' \leq t} \dot \phi_i^{t'} \delta t + \phi_i^0$ using a $\delta t = 1\,\mathrm{ms}$. The initial values were set to $\phi_1^0 = 0$ and $\phi_2^0 = \frac{\pi}{2}$.

Feasible target movements $\bm y^{*,t}$ of duration $500$ ms were generated randomly by sampling the generating angular velocities $\dot \Phi^{*,t}=(\dot \phi^{*,t}_1, \dot \phi^{*,t}_2)$. Each of the target angular velocities exhibitted a common form

\begin{eqnarray}
	\dot \phi^{*,t}_i = \sum_m S_{im} \sin \left( 2 \pi \omega_{im} \frac{t}{T} + \delta_{im} \right) \stackrel{\mathrm{def}}{=} \sum_m q_{im}^t~,
\end{eqnarray}
where the number of components $m$ was set to 5, $S_{im}$ was sampled uniformly in $[0, 30]$, $\omega_{im}$ was sampled uniformly in $[0.3, 1]$ and $\delta_{im}$ was sampled uniformly in $[0, 2 \pi]$. After this sampling, every component $q_{2,m}^t$ in $\dot \phi^{*,t}_2$ was rescaled to satisfy $\max_t(q_{2,m}^t) - \min_t(q_{2,m}^t) = 20$. In addition, we considered constraints on the angles of the joints: $\phi_1 \in [- \frac{\pi}{2}, \frac{\pi}{2}]$ and $\phi_2 \in [0, \pi]$. If violated, the respective motor commands $\dot \phi_i^{*,t}$ were rescaled to match the constraints.

A clock-like input signal was implemented as in task 1.1 by 20 input neurons, that fired in groups in 5 successive time steps with a length of 100 ms at a rate of 100 Hz.

\textbf{Outer loop optimization:}
The procedure described above defines an infinitely large familiy of tasks, each task of the family being one particular target movement. We optimized the parameters of the error module as well as the initial parameters of the learning network in an outer-loop optimization procedure.
The learning cost $\mathcal{L}_C$ for tasks $C$ in the above defined family of tasks was defined as 
\begin{equation}
	\mathcal{L}_C(\bm \theta_{\mathrm{test}, C}) = \sum_t \left( \left( \bm y^t(\bm \theta_{\mathrm{test}, C}) - \bm y^{*,t} \right)^2 + \left( \dot \Phi^t(\bm \theta_{\mathrm{test}, C}) - \dot \Phi^{*,t} \right)^2 \right) + \lambda E_\mathrm{reg} \label{eq:e-prop2-ltl-objective}
\end{equation}
to measure how well the target movement was reproduced. We then optimized the expected cost over the family of learning task using BPTT. In addition, a regularization term for firing rates as defined in equation~\eqref{eq:regularization} was introduced with $\lambda = 0.25$. Gradients were computed over batches of 200 different tasks to empirically estimate the learning cost across the family of tasks: ${\mathds{E}_{C\sim \mathcal{F}} \left[ \mathcal{L}_C(\bm \theta_{\mathrm{test},C}) \right] \approx \frac{1}{200} \sum_{i =1}^{200} \mathcal{L}_{C_i}(\bm \theta_{\mathrm{test},C_i})}$. We used the Adam algorithm~\citep{kingma2014adam} with a learning rate of $0.0015$. The learning rate decayed after every $300$ steps by a factor of $0.95$.

\textbf{Model parameters:}
The learning network consisted of 400 LIF neurons according to the model stated in equation~\eqref{eq:lifv} and~\eqref{eq:lifz}, with a membrane time constant of $20$ ms and a threshold of $v_\mathrm{th} = 0.4$. The motor commands $\dot \phi_j^t$ predicted by the network were given by the output of readout neurons with a membrane time constant of $20$ ms. The target firing rate in the regularizer was set to $f^{target} = 20\,\mathrm{Hz}$. 

The error module was implemented as a recurrently connected network of 300 LIF neurons, which had the same membrane decay as the learning network. The neurons in the error module were set to have a threshold of $v_\mathrm{th} = 0.4$. Readout neurons of the error module had a membrane time constant of $20$ ms. Finally, the weight update with \textit{e-prop} according to equation~\eqref{eq:ltl_weight_update} used a learning rate of $\eta = 10^{-4}$.  The target firing rate in the regularizer was set to $f^{target} = 10\,\mathrm{Hz}$.

Both the learning network as well as the error module used a refractory period of $5\,$ms.

\textbf{Linear error module:}
The alternative implementation of a linear error module was implemented as a linear mapping of inputs formerly received by the spiking implementation of the error module. Prior to the linear mapping, we applied a filter to the spiking quantities $\bm x^t$, $\bm z^t$ such that $\hat{\bm x}^t = \sum_{t' \leq t} \alpha_e^{t - t'} \bm x^t$ and similarly for $\hat{ \bm z}^t$. Then, the learning signal from the linear error module was given as: $\hat L_j^t = \sum_i \Phi_{ji}^x \hat x_i^t + \sum_i \Phi_{ji}^z \hat z_i^t + \sum_i \Phi_{ji}^y y_i^{*,t}$

\subsection*{\textit{E-prop 3}}
We first describe \textit{e-prop 3} in theoretical terms when the simulation duration is split into intervals of length $\Delta t$ and show two mathematical properties of the algorithm: first, it computes the correct error gradients if the synthetic gradients are ideal; and second, when the synthetic gradients are imperfect, the estimated gradients are a better approximation of the true error gradient in comparison to BPTT.
In subsequent paragraphs we discuss details of the implementation of \textit{e-prop 3}, the computation of the synthetic gradients and hyperparameters used in tasks 3.1 and 3.2.

\paragraph{Notation and review of truncated BPTT:}
We consider the true error gradient $\frac{dE}{d \theta_{ji}}$ to be the error gradient computed over the full simulation ranging from time $t=1$ to time $T$. Truncated BPTT computes an approximation 
of this gradient. In this paragraph, we identify the approximations induced by truncated BPTT. 

In truncated BPTT, the network simulation is divided into $K$ successive intervals of length $\Delta t$ each. For simplicity we assume that $T$ is a multiple of $\Delta t$, such that $K=T/\Delta t$ is an integer. 
Using the shorthand notation $t_m=m \Delta t$, the simulation intervals are thus $\{1, \dots, t_1\}, \{t_1+1, \dots, t_2\},\dots, \{t_{K-1}+1, \dots, t_K\}$.
To simplify the theory we assume that updates are implemented after the processing of all these intervals (i.e., after time $T$).

For each interval $\{t_{m-1}+1,\dots, t_m\}$, the simulation is initialized with the network state $\bt s^{t_{m-1}}$. Then, the observable states $\bt z^{t'}$ and hidden states $\bt s^{t'}$ are computed for $t' \in \{ t_{m-1}+1, \dots, t_m \}$.
It is common to use for the overall error $E(\bt z_1, \dots, \bt z_T)$ an error function that is given by the sum of errors in each individual time step. Hence the error can be written as a sum of errors $E_{m}(\bt z^{t_{m-1} + 1}, \dots, \bt z^{t_{m}})$ in the intervals:
\begin{equation}
E(\bt z^1, \dots, \bt z^T) = \sum_{m=1}^K E_{m}(\bt z^{t_{m-1} + 1}, \dots, \bt z^{t_{m}})~.
\end{equation}
For each such interval, after network simulation until $t_{m}$ (the forward pass), the gradients $\frac{dE_{m}}{d\bt s_j^{t'}}$ are propagated backward from $t'=t_{m}$ to $t'=t_{m-1}  + 1$ (the backward pass). The contribution to the error gradient for some paramter $\theta_{ji}$ in the interval is then given by (compare to equation \eqref{eq:bptt})
\begin{eqnarray}
 g_{m,ji}^{\mathrm{trunc}}  & = & \sum_{t'=t_{m-1} + 1}^{t_{m}} \frac{dE_m}{d \bt s_j^{t'}} \cdot \frac{\partial \bt s_j^{t'}}{\partial \theta_{ji}}~.  \label{eq:truncated-bptt}
\end{eqnarray}
The overall gradient $\frac{dE}{d \theta_{ji}}$ is then approximated by the sum of the gradients in the intervals:
$g_{1,ji}^{\mathrm{trunc}} + g_{2,ji}^{\mathrm{trunc}} + \dots + g_{K,ji}^{\mathrm{trunc}}$.

This approximation is in general not equal to the true error gradient $\frac{dE}{d \theta_{ji}}$, as it disregards the contributions of network outputs within one interval on errors that occur in a later interval. 

\textbf{Synthetic gradients:}
To correct for the truncated gradient, one can provide a suitable boundary condition at the end of each interval that supplements the missing gradient.
The optimal boundary condition 
cannot be computed in an online manner since it depends on future activities and future errors that are not yet available. In truncated BPTT, one chooses 
$\frac{dE}{d\bt s_j^{t_m + 1}}=0$ at the end of an interval $\{t_{m-1}+1 ,\dots, t_m\}$, which is exact only if the simulation terminates at time $t_{m}$ or if future errors do not depend on network states of this interval.
The role of synthetic gradients is to correct this approximation by providing a black box boundary condition $\SG_j(\bt z^{t_m}, \Psi)$, where $\SG_j$ is a parameterized function of the network output with parameters $\Psi$. $\SG_j$ should approximate the optimal boundary condition, i.e., ${\SG_j(\bt z^{t_m}, \Psi) \approx \frac{dE}{d\bt s_j^{t_m+1}}}$.


We denote the approximate gradient that includes the boundary condition given by synthetic gradients by $\frac{d \overline E}{d\bt s_j^{t}}$. This gradient is given by 
\begin{equation}
{\frac{d \overline{E}_m}{d \bm s_j^t} = \frac{d E_m}{d \bm s_j^t} + \eta_{SG} \sum_l \SG_l(\bm z^{t_m}, \Psi) \frac{d \bm s_l^{t_m +1}}{d \bm s_j^t} }.\label{eq:grad_sga}
\end{equation}
We will continue our theoretical analysis with a factor $\eta_{SG}=1$ (as suggested in \cite{jaderberg2016decoupled}, we set  $\eta_{SG}$ to $0.1$ in simulations to stabilize learning).
We define $g_{m,ji}^\mathrm{SG}$ as the corrected version of $g_{m,ji}^\mathrm{trunc}$ that incorporates the new boundary condition. We finally define the estimator of the error gradient with synthetic gradients as:
\begin{eqnarray}
	\widehat{ \frac{dE}{d \theta_{ji}} }^{\mathrm{SG}} & = &
	g_{1,ji}^{\mathrm{SG}} + g_{2.ji}^{\mathrm{SG}} + \dots + g_{K,ji}^{\mathrm{SG}}. \label{eq:grad-truncated-bptt}
\end{eqnarray}
The synthetic gradient approximation is refined by minimizing the mean squared error between the synthetic gradient approximation $\SG_j(\bt z^{t_m}, \Psi)$ and the gradient $\frac{d \overline{E}_{m + 1}}{d \bt s_j^{t_m+1}}$, which is computed in the interval $t_{m}+1$ to $t_{m +1}$ and includes the next boundary condition $\SG_l(\bt z^{t_{m + 1}})$:
\begin{eqnarray}
	E_{\SG} \left( \bt z^{t_m},{\frac{d \overline{E}_{m+1}}{d \bt s_j^{t_m+1}}}, \Psi \right) & = & \sum_j \frac{1}{2} \norm{ SG_j(\bt z^{t_m}, \Psi) - {\frac{d\overline E_{m+1}}{d \bt s_j^{t_m+1}}} }^2~. \label{eq:sg-loss}
\end{eqnarray}
\textbf{Correctness of synthetic gradients:}
We consider $\Psi^*$ to be optimal synthetic gradient parameters if the synthetic gradient loss in equation \eqref{eq:sg-loss} is always zero.
In this case, all synthetic gradients $\SG(\bt z^{t_{m}}, \Psi^*)$ exactly match $\frac{dE}{d \bt s_j^{t_{m}+1}}$, and the computed approximation exactly matches the true gradient.
This analysis assumes the existence of the optimal parameters $\Psi^*$ and the convergence of the optimization algorithm to the optimal parameters. This is not necessarily true in practice. For an analysis of the convergence of the optimization of the synthetic gradient loss we refer to \citep{czarnecki2017understanding}.

\paragraph{Proof of correctness of \textit{e-prop 3} with truncated time intervals:}
Similarly to the justification above for synthetic gradients, we show now that the error gradients $\frac{d E}{d \theta_{ji}}$ can be estimated with \textit{e-prop 3} when the gradients are computed over truncated intervals.

Instead of using the factorization of the error gradients as in BPTT (equation \eqref{eq:bptt}), \textit{e-prop 3} uses equation \eqref{eq:grad}.
The approximate gradient that is computed by \textit{e-prop 3} with respect to neuron outputs is given analogously to equation \eqref{eq:grad_sga}
\begin{equation}
{\frac{d \overline{E}_m}{d z_j^t} = \frac{d E_m}{d z_j^t} + \sum_l \SG_l(\bm z^{t_m}, \Psi) \frac{d \bm s_l^{t_m +1}}{d z_j^t} }.
\end{equation}\label{eq:grad_mergea}
We are defining the learning signal as in equation \eqref{eq:L}, but now using the enhanced estimate of the derivative of the interval error:
\begin{equation}
 \overline{\bm L}_{m,j}^{t}=\frac{d \overline{E}_m}{dz_j^{t}} \frac{dz_j^{t}}{d \bt s_j^{t}}.
\end{equation}
This learning signal is computed recursively using equation \eqref{eq:backpropagation} within an interval. At the upper boundaries $t_m$ of the intervals, the boundary condition is computed via synthetic gradients.

Analogous to $g_{m,ji}^{\mathrm{SG}}$, we define the gradient approximation of \textit{e-prop 3} $g_{m,ji}^{\mathrm{e-prop}}$ as the corrected version of $g_{m,ji}^\mathrm{trunc}$ that incorporates the boundary condition for interval $\{t_{m-1}+1, \dots, t_{m}\}$ via synthetic gradients. This gradient approximation is given by \begin{equation}
g_{m,ji}^{\mathrm{e-prop}} = \sum_{t=t_{m-1} + 1}^{t_{m}} \overline{L}_{m,j}^{t} \cdot \bt \bepsi_{ji}^{t}~.
\end{equation}
Considering the sum of terms $g_{m,ji}^{\mathrm{e-prop}}$ associated with each interval, we write the estimator of the true error gradient computed with \textit{e-prop 3} as:
\begin{eqnarray}
	\widehat{\frac{dE}{d \theta_{ji}}}^{\mathrm{e-prop}} & = &
	g_{1,ji}^{\mathrm{e-prop}} +
	g_{2,ji}^{\mathrm{e-prop}} +
	\dots + 
	g_{K,ji}^{\mathrm{e-prop}}~. \label{eq:e-prop-estimator}
\end{eqnarray}
Assuming now that this boundary condition is provided by an error module computing the synthetic gradients $\SG$ with optimal parameters $\Psi^*$. 
As explained above, it follows that all $\SG_l(\bt z^{t_m}, \Psi^*)$ computes exactly $\frac{dE}{d \bt s_l^{t_m+1}}$ which is true independently of the usage of BPTT or \textit{e-prop 3}. In the later case, it follows that $\frac{d \overline{E}_m}{dz_j^{t}}$ is correctly computing $\frac{d E}{dz_j^{t}}$ and hence, $\overline{\bm L}_{m,j}^{t}$ is equal to the true learning signal $\bm L_{j}^{t}$.
Looking back at equation \eqref{eq:grad}, it follows that the estimator defined at equation \eqref{eq:e-prop-estimator} is equal to the true gradient if the parameters of the error module are optimal.

\paragraph{Optimization of the synthetic gradient parameters $\Psi$:}
We define here the algorithm used to optimize the synthetic gradients parameters $\Psi$ and the network parameters $\bth$. Using the same truncation scheme as described previously, we recall that the loss function $\overline{E}_m$ formalizes the loss function on interval $m$ denoted $E_m$ with the modification that it takes into account the boundary condition defined by the synthetic gradients. We then consider the loss $E'$ as the sum of the term $\overline{E}_m$ and the synthetic gradient loss $E_{SG}$.
The final algorithm is summarized by the pseudo-code given in Algorithm \ref{alg:sg}.
Note that this algorithm is slightly different from the one used originally by \cite{jaderberg2016decoupled}.
Our version requires one extra pair of forward and backward passes on each truncated interval but we found it easier to implement.

\begin{algorithm}
\For{$m \in \{1, \dots, K\} $}
{
 Simulate the network over the interval $\{t_{m - 1} + 1,\dots, t_m\}$ to compute the network states $\bt s_j^t$ \\
 Backpropagate gradients on the interval $\{t_{m - 1} + 1,\dots, t_m\}$ to compute $\frac{d \overline{E}_m}{d \bth}$ using the boundary condition provided by $\SG_l(\bm z^{t_m}, \Psi)$. Store $\bt s_j^{t_m}$ and $\bt \bepsi_{ji}^{t_m}$ to be used as initial states in the next interval.\\
 Simulate the network over the interval $\{t_{m} + 1,\dots, t_{m + 1}\}$ to compute the network states $\bt s_j^t$ \\
 Backpropagate gradients on the interval $\{t_{m} + 1,\dots, t_{m + 1}\}$ to obtain ${\frac{d \overline{E}_{m+1}}{d \bt s_j^{t_m+1}}}$ and compute $\frac{d E_{\SG}}{d \bth}$, $\frac{d E_{\SG}}{d \Psi}$, \\
 Update the parameters $\Psi$ and $\bth$ using $\frac{d (E_m + E_{\SG})}{d \bth}$ and $\frac{d E_{\SG}}{d \Psi}$ with any variant of stochastic gradient descent
}
\caption{Pseudo code to describe the algorithm used to trained simultaneously the network parameters $\bth$ and the synthetic gradients $\Psi$ in both \textit{e-prop 3} and BPTT with synthetic gradients.} \label{alg:sg}
\end{algorithm}

\paragraph{Copy-repeat task 3.1:}
Each sequence of the input of the copy repeat task consists of the ``8-bit'' pattern of length $n_{\mathrm{pattern}}$ encoded by 8 binary inputs, a stop character encoded by a 9th binary input channel, and a number of repetitions $n_{\mathrm{repetitions}}$  encoded using a one hot encoding over the $9$ input channels. 
While the input is provided, no output target is defined. After that the input becomes silent and the output target is defined by the $n_{\mathrm{repetitions}}$ copies of the input followed by a stop character. As for the input, the output pattern is encoded with the first $8$ output channels and the 9-th channel is used for the stop character.
Denoting the target output $b_k^{*,t}$ of the channel $k$ at time $t$ and defining $\sigma(y_k^{t})$ as the output of the network with $y_k^{t}$ a weighted sum of the observable states $z_j^t$ and $\sigma$ the sigmoid function, the loss function is defined by the binary cross-entropy loss: $E = - \sum_{t,k} (1- b_k^{*,t}) \operatorname{log}_2{\sigma(y_k^{t})} + b_k^{*,t} \operatorname{log}_2{\left(1 - \sigma(y_j^{t})\right)}$.
The sum is running over the time steps where the output is specified.

We follow the curriculum of \cite{jaderberg2016decoupled} to increase gradually the complexity of the task: when the error $E$ averaged over a batch of $256$ sequences is below $0.15$ bits per sequences, $n_{\mathrm{pattern}}$  or $n_{\mathrm{repetitions}}$ are incremented by one.
When the experiments begins, we initialize $n_{\mathrm{pattern}}$ and $n_{\mathrm{repetitions}}$ to one. After the first threshold crossing $n_{\mathrm{pattern}}$ is incremented, then the increments are alternating between $n_{\mathrm{pattern}}$ and $n_{\mathrm{repetitions}}$.

For each batch of $256$ sequences, the parameters are updated every $\Delta t=4$ time steps when the simulation duration in truncated as in BPTT.
The parameter updates are applied with Adam, using learning rate $0.0001$ and the default hyperparameters suggested by \cite{kingma2014adam}.

\paragraph{Word prediction task 3.2:}
Training was performed for $20$ epochs, where one epoch denotes a single pass through the complete dataset. All learning rules used gradient descent to minimize loss with initial learning rate of $1$ which was decayed after every epoch with factor $0.5$, starting with epoch $5$.
Mini-batch consisted of $20$ sequences of length $\Delta t$.
Sequence of sentences in Penn Treebank dataset are connected and coherent, so the network state was reset only after every epoch.
Equally the eligibility traces are set to zero at the beginning of every epoch.

\subsection*{Acknowledgments}

This research was supported by the Human Brain Project of the European Union, Grant agreement No. 785907.
We also gratefully acknowledge the support of NVIDIA Corporation with the donation of the Quadro P6000 GPU used for this research.
Computations were primarily carried out on the Supercomputer JUWELS at Jülich Supercomputing Centre.
We gratefully acknowledge the support of the SimLab of the Forschungszentrum Jülich in securing the grant CHHD34 ``Learning to Learn on Spiking Neural Networks''
from the Gauss Centre for Supercomputing, which provided funding for this computing time.
We also gratefully acknowledge the Vienna Scientific Cluster (VSC) for providing additional computing time.
We would like to thank Arjun Rao for his contribution to the software used in our experiments.
We also want to thank him, Mike Davies, Michael M\"uller, Christoph Stoeckl, and Anand Subramoney for comments on earlier version of the manuscript.

\bibliographystyle{apalike}
\bibliography{library}

\end{document}